\documentclass[nonacm, acmsmall]{acmart}

\AtBeginDocument{%
  }

\usepackage{amsmath}
\usepackage{balance}       
\usepackage{graphics}      
\usepackage[T1]{fontenc}   
\usepackage{lipsum}
\usepackage{color}
\usepackage{colortbl}
\usepackage{booktabs}
\usepackage{textcomp}
\usepackage{microtype}        
\usepackage{ccicons}          
\usepackage{multirow}  
\usepackage{color,soul}  
\usepackage{subcaption}  
\usepackage{listings}  
\usepackage{todonotes}
\usepackage{multirow}

\usepackage{enumitem} 

\usepackage{array}
\newcolumntype{P}[1]{>{\centering\arraybackslash}p{#1}}

\definecolor{ao(english)}{rgb}{0.0, 0.5, 0.0}

\newcommand{\edit}[1]{{\textcolor{black}{#1}}}

\def\etal{\emph{et al.\ }}

\let\oldsim\sim 
\renewcommand{\sim}{{\oldsim}}

\begin{document}

\title{Cross-Domain HAR: Few Shot Transfer Learning for Human Activity Recognition }

\author{Megha Thukral}
\email{mthukral3@gatech.edu}
\affiliation{%
	\institution{School of Interactive Computing, Georgia Institute of Technology}
	\city{Atlanta, GA}
	\country{USA}
}

\author{Harish Haresamudram}
\email{hharesamudram3@gatech.edu}
\affiliation{%
	\institution{School of Electrical and Computer Engineering, Georgia Institute of Technology}
	\city{Atlanta, GA}
	\country{USA}
}

\author{Thomas Pl\"{o}tz}
\email{thomas.ploetz@gatech.edu}
\affiliation{
	\institution{School of Interactive Computing, Georgia Institute of Technology}
	\city{Atlanta, GA}
	\country{USA}
}

\renewcommand{\shortauthors}{Thukral et al.}

\begin{abstract}
   The ubiquitous availability of smartphones and smartwatches with integrated inertial measurement units (IMUs) enables  straightforward capturing of human activities through collecting movement data. 
For specific applications of sensor based human activity recognition (HAR), however, logistical challenges and burgeoning costs render especially the ground truth annotation of such data a difficult endeavor, resulting in limited scale and diversity of datasets available for deriving effective HAR systems and less than ideal recognition capabilities. 
Transfer learning, i.e., leveraging publicly available labeled datasets to first learn useful representations that can then be fine-tuned using limited amounts of labeled data from a target domain, can alleviate some of the performance issues of contemporary HAR systems.
Yet they can fail when the differences between source and target conditions are too large \edit{and / or only few samples from a target application domain are available -- each of which are typical challenges in real-world human activity recognition scenarios}. 
In this paper, we present an approach for economic use of publicly available labeled HAR datasets for effective transfer learning.
We introduce a novel transfer learning framework--Cross-Domain HAR--which follows the teacher-student self-training paradigm to more effectively recognize activities with very limited label information.
It bridges conceptual gaps between source and target domains, including sensor locations and type of activities. 
Cross-Domain HAR enables substantial performance improvements over the state-of-the-art in sensor-based HAR scenarios.
Through our extensive experimental evaluation on a range of benchmark datasets we specifically demonstrate the effectiveness of our approach for practically relevant few shot activity recognition scenarios. 
We also present a detailed analysis into how the individual components of our framework affect downstream performance and provide practical suggestions for using the framework in real-world applications.
\end{abstract}




\maketitle

\section{Introduction}

The ubiquitous availability of movement sensors (specifically: Inertial Measurement Units -- IMUs) on everyday smartphones and smartwatches enables the collection of large-scale time-series human movement data.
Human Activity Recognition (HAR), which utilizes such movement sensors to recognize activities being performed, has far reaching impact  in various application areas including fitness tracking, manufacturing monitoring, healthcare and well-being assessments \cite{bachlin2009wearable, morshed2019prediction, banos2014mhealthdroid,zappi2012network, bin2020measuring}.
Deep learning--especially supervised learning--has enabled significant progress towards developing (and deploying) accurate and reliable recognition systems \cite{hammerla2016deep, morales2016deep}.
Traditional supervised learning methods, however, rely heavily on large amounts of labeled, raw sensor data; yet annotating these  is a difficult proposition and requires significant effort \cite{kwon2020imutube}. 
The lack of substantial, large-scale labeled datasets, coupled with the relative ease of collecting  unlabeled sensor data has led to the adoption of  data-efficient methods involving self-supervised and  semi-supervised learning.  

Self-supervised learning, an unsupervised learning method \cite{ericsson2022self}, leverages large amounts of unlabeled data to learn rich representations by devising pretext tasks from the data itself \cite{haresamudram2021contrastive, haresamudram2022assessing}, i.e., without any annotations.
These pretext tasks assist in generating supervisory signals necessary for pre-training models to capture semantic features from raw data.
In HAR,  several self-supervised methods such as Multi-task self-supervision \cite{saeed2019multi}, Contrastive Predictive Coding (CPC) \cite{haresamudram2021contrastive}, and SimCLR \cite{tang2020exploring} have been successfully deployed to improve  recognition performance thereby minimizing  annotation cost and effort.
The success of self-supervised methods for building wearable sensor based recognition systems has motivated the deployment of other such  methods, which can further reduce reliance on annotations and leverage existing labeled/ unlabeled datasets.
While self-supervised methods utilize significant amounts of unlabeled data to overcome the lack of annotations, semi-supervised methods\cite{hady2013semi, ouali2020overview} such as self-training and transfer learning can utilize  existing labeled data as well as unlabeled sensor streams. 

In application scenarios where only a limited amount of sensor data can be collected and annotated, leveraging publicly available, labeled data can lead to performance improvements for many real-world HAR applications.
However, such approaches do not come without challenges.
In most application scenarios, data collection is performed by recruiting only a few users, who are often asked to conduct orchestrated sets of activities of daily living  (most commonly sitting, standing and walking), while positioning sensors on locations such as wrist (e.g., smartwatches) and trouser pockets (e.g., via mobile phones). 
Utilizing such existing, annotated datasets, however, becomes a difficult task, when target applications cover  different sets of activities and sensor placements.
For example, if we aim to build a recognition system for covering certain activities in the healthcare domain, a specific set of  movements and activities  needs to  be collected, which may not be covered by common activities like sitting, standing and walking, i.e., by existing datasets. 
In addition, the target application may require sensor placements that are different than  those used in the publicly available labeled datasets.

To effectively utilize  well-curated and annotated, publicly available HAR datasets, we present a transfer learning \cite{cook2013transfer} method that can bridge the large domain gap between application domains.
Developing such a method can assist in minimizing the substantial cost, effort and domain expertise needed for annotations and can be instrumental in bootstrapping activity recognition models quickly for real-world applications.

Our method is based on teacher-student based self-training to adapt feature representations learned from  labeled source domains  to  target domains thereby requiring only a few seconds of labeled data.
We add  consistency regularization \cite{ouali2020overview, islam2021dynamic} by minimizing the distance between teacher model predictions for  weakly augmented unlabeled target sensor signals, and student model predictions for the strongly augmented version of the same input signal. 
We apply eight different augmentations to the source data to  artificially increase the size of available--much smaller--activity recognition datasets.
Our method  filters out noisy predictions for target data based on the teacher model confidence and also incorporates self-supervised loss on unlabeled target data  within the standard student model training. 
Our  self-training process is deployed to effectively bridge the  gap between source and target  domains, with the goal to derive robust sensor based recognition systems for new application domains.

The contributions of this paper can be summarized as follows:
\begin{enumerate}
\item We present a transfer learning method--Cross-Domain HAR--which utilizes a teacher-student based self training process  to build recognition systems with very limited labeled data availability. 
Our method incorporates  self-supervision, consistency regularization, and source data augmentation within the self training framework to improve few shot evaluation for target domains. 
Our method uses both labeled source dataset and unlabeled target dataset to adapt representations from source  to target domains.  
\item We evaluate our method on six diverse datasets containing various activities recorded from different on-body sensor placements. 
We demonstrate that our method achieves significant performance improvements in terms of activity recognition when compared to the state-of-the-art.
\item We conduct extensive experimentation to analyse the components of our method to determine the conditions impacting successful transfer and show how they impact downstream recognition performance.
\end{enumerate}

\vspace*{-0.5em}
\section{Related Work}
\label{sec:related}
In our work, we focus on learning effective representations for movement data collected using on-body sensors for  application domains which have  only a few seconds of labeled sensor data per activity of interest available. 
For this purpose, we aim to utilize unlabeled data from a target domain and  existing annotated activity data from a different domain/ application covering a distinct group of activities, users, and on-body sensor positions. 
To transfer knowledge in such a few shot setup, we follow the self-training based teacher-student learning paradigm and also incorporate self-supervision on target, unlabeled data.
Our work is in line with efforts that aim at reducing reliance on large amounts of annotated movement data to learn rich representations and build accurate classification models for activity recognition.

Relevant prior work involves the following: (i) self-training in general; (ii) self-supervised learning methods for HAR;  (iii) transfer learning for HAR; and (iv) few shot learning in general.

\subsection{Self-Training } 
Self training is a widely used technique in semi-supervised learning as well as knowledge distillation \cite{hinton2015distilling, phoo2020self}.
It comprises three steps: 
\emph{a)} training a teacher model on labeled data; 
\emph{b)} generating pseudo-labels for unlabeled data using the teacher model; and 
\emph{c)} training a new student model using both labeled and unlabeled data \cite{tang2021selfhar}.

In other domains, such as computer vision, this method has been used effectively for improving, e.g., object classification (ImageNet) \cite{deng2009imagenet} performance. 
\cite{yalniz2019billion} deploys the self-training teacher/student paradigm to  leverage a collection of up to one billion unlabelled images. 
In their method, the authors use the top-K confident samples per class as they are predicted by a teacher model to train a   student model. 
Significant performance improvements for image classification using self-training were achieved.
\cite{xie2020self} used a noisy student model in the standard self-training framework. 
Adding different noise inputs, such as  data augmentation, dropout and stochastic depth, encouraged the student
model to more generalized features and learn more than the teacher. 
Utilizing noise in a modified self-training approach resulted in performance improvements for ImageNet data.

STARTUP \cite{phoo2020self} used the self-training approach for few shot transfer learning across extremely different tasks and domains. 
The authors studied how  few shot transfer learning can be improved when the source dataset is ImageNet while the target domains/datasets contain satellite images, chest x-rays, or images of crop diseases, i.e., \textit{very} different domains.
Similarly, Islam \etal \cite{islam2021dynamic} used the dynamic distillation-based self-training approach to perform cross domain few shot learning. The authors deploy a mean-teacher approach where teacher model weights are updated as the exponential moving average of the parameters of a student model.
This method also imposes consistency regularization by  matching the  predictions for the weakly-augmented versions of the unlabeled images from a teacher model with student network output for the strongly augmented versions of the same images.
Our transfer learning framework for sensor-based human activity recognition applies  modifications to self-training process similar to \cite{phoo2020self} and \cite{islam2021dynamic}, with the goal of transferring knowledge from an  existing domain to  different target domains.

In HAR, Tang \textit{et al.} \cite{tang2021selfhar} introduced Self-HAR which combined the teacher-student self-training and multi-task self-supervised learning for leveraging unlabeled data for improving activity recognition performance. 
Their approach consists of the following steps: 
(1) train a teacher model on target labeled data; 
(2) generate self-training labels for the unlabeled and labeled data using the teacher model; 
(3) filter out the most confident top-K predictions as generated by the teacher model, and augment them; 
(4) pre-train a student model in a multi-task learning setting using filtered augmented data; and 
(5) fine-tune the student model with the target labeled dataset. 

The goal of Self-HAR is to improve the performance for a target domain where a labeled dataset is already available. 
In contrast, we aim to leverage an existing labeled dataset to annotate the unlabeled dataset from a completely different domain. 
Our strategy aims to bridge the  domain gap by transferring the knowledge gained from a source domain to build a new classifier for the novel target domain with unseen activities and different sensor positions -- a scenario that is of substantial relevance for sensor-based HAR but has not been explored yet.

\subsection{Self-Supervised Learning} 
Self-supervised learning is a form of unsupervised learning, which derives supervisory signals for downstream performance by creating auxiliary tasks \cite{haresamudram2022assessing,  saeed2019multi, jain2022collossl} on unlabeled data. 
The models are pre-trained on these pretext/auxiliary tasks to steer learning towards extracting more generalized features from the data itself.
In self-supervision, a model learns in two steps: 
\emph{i)} pre-training -- model parameters are learned by solving the pretext task on the unlabeled data; and 
\emph{ii)} fine-tuning -- where the learned parameters function as the feature extractor towards the actual downstream task \cite{haresamudram2021contrastive}.

For HAR, a multitude of pretext tasks has been deployed for learning effective representations from time-series movement data. 
Saeed \etal \cite{saeed2019multi} probabilistically applied augmentations to sensor data, and predicted whether they had been applied or not, in a multi-task setting.  
Random parts of the input signals are masked/perturbed in \cite{haresamudram2020masked}, and the model is forced  to reconstruct only the masked out parts, thereby learning local temporal patterns from the surrounding context.
In contrast, Contrastive Predictive Coding (CPC) utilizes long term prediction of the future timesteps in a contrastive setting as a pretext task \cite{haresamudram2021contrastive}, with the goal of capturing the underlying long-term signal of the data.
SimCLR \cite{chen2020simple} has been adapted for sensor-based HAR, where two transformations (out of a total of eight possibilities) are applied to each window, resulting in positive pairs, whereas all other pairs comprise the negatives/distractors for contrastive learning \cite{tang2020exploring}. 
NT-Xent (normalized temperature-scaled cross entropy) loss is applied to maximize the agreement between differently augmented versions of the same input.
Other augmentation based approaches such as SimSiam and BYOL have also been adopted and applied to wearables applications and are typically highly effective.

Self-supervised learning methods have shown great promise for leveraging unlabeled datasets; however these methods require large amounts  of data to achieve  performance comparable to supervised learning methods \cite{saeed2019multi, haresamudram2021contrastive}. 
Moreover, these methods start \textit{tabula rasa}, i.e., they learn everything about the data from scratch \cite{phoo2020self}.
On the other hand, given access to a labeled dataset, self-training based semi-supervised method can leverage both target unlabeled dataset as well as source labeled data. 
The existing knowledge can be effectively transferred  to bootstrap models in a new domain using self-training approach.

In Cross-Domain HAR, we utilize self-supervised learning in addition to the self-training process, specifically SimCLR self-supervision (based on \cite{tang2020exploring}) on unlabeled target data, as in STARTUP \cite{phoo2020self}.
The addition of the SimCLR-based loss is intended to assist the student model for extracting additional knowledge specific to the target domain, albeit in a self-supervised setting.

\vspace*{-1em}
\subsection{Transfer Learning} 
When deployed in the real world, the performance of sensor-based HAR models can be negatively impacted due to differences between the training data and testing conditions.
These differences can be attributed to variations in how users perform activities, device heterogeneity, sensor placement, the physical environment, and presence of new/unseen activity classes \cite{cook2013transfer}.
With the resulting changes to the data distributions between training and test data, degraded performance may be observed for deep learning methods, which  necessitates retraining models with large amounts of annotated activity data for every new application. 

The high costs and effort in curating and annotating large-scale data have encouraged researchers to deploy other, more data efficient techniques of model building such as  transfer learning. 
It has been observed that for sensor-based HAR, supervised transfer learning is generally challenging. 
For example, \cite{hoelzemann2020digging} conduct an analysis of the transfer learning performance between PAMAP2 \cite{reiss2012introducing} and Skoda Mini checkpoint \cite{zappi2012network}. 
A significant drop in performance is observed when the model weights are transferred for inference on the other dataset. 
Such a reduction is seen despite the datasets having the same sensor positions and modalities, while only activities differ.
Therefore, the lower performance shows a lack of generalizability of the features learned from one dataset for application to another task/application.

Another paper \cite{ding2018empirical} studied transfer learning across users within the same datasets of UCI-HAR \cite{altun2010comparative} and USC-HAD \cite{zhang2012usc}, thereby keeping the sensor modalities and activities constant.
It was found that the representations learned for each user have high intra-class and low inter-class variances, rendering the learning of generalized features a challenging endeavor.
Maximum Mean Discrepancy (MMD) was used to perform feature representation transfer across users. 
Similarly, An \etal \cite{an2020transfer} performed transfer learning across users so as to analyze, which layers of a network capture the most common features across user clusters. 
The authors observed that only the first few layers are somewhat useful for transfer because later layers captured only task specific features. 

Morales \etal \cite{morales2016deep} also found that transferring across datasets that cover different activities  leads to significant performance degradation, even when the on-body sensor locations are the same. 
Instead, transferring across sensor placements or across sensor modalities within the same dataset (keeping the application domain the same) seems to be a more suitable scenario for transferring knowledge across models.

These studies clearly indicate  the challenging nature of transfer learning in sensor-based HAR, especially across application domains.
In our work, we tackle the difficult problem to bridge the domain gap between source and target application scenarios where users, on-body sensor locations, as well as activities are different. 
Furthermore, we study situations where there are only few seconds of labeled target data per class available for adapting the model to target conditions, thereby increasing the complexity of the task.

\vspace*{-1em}
\subsection{Few Shot Learning}
Few shot Learning (FSL) \cite{wang2020generalizing} is a machine learning paradigm where models learn from a limited number of examples with supervisory signals for the target task.
In recent years, FSL has gained popularity as it not only leads to  reduction in data collection efforts and annotation costs but also  represents the next step in moving towards human like intelligence.
For example, if given access to only two or three images of a new kind of object, a child is able to transfer their knowledge about concepts of similarity or dissimilarity to identify/distinguish  this new object from a larger collection of images.
In a similar manner, for solving few shot learning, recognition models need to utilize prior knowledge (gained in some other task or domain) to generalize and perform well on  new tasks rapidly using an extremely low number of labeled data points.

FSL has been applied for many applications and tasks like image classification, language modelling and reinforcement learning\cite{wang2020generalizing}.
 \cite{vinyals2016matching} applies a matching network architecture to learn a mapping between few samples, its corresponding unlabelled example and  class label to achieve one-shot accuracy improvement for Imagenet \cite{deng2009imagenet} data and language modelling tasks.
\cite{snell2017prototypical} deploys prototypical networks, which classify a query image by computing its distance to the prototype of each class's  embedding cluster and evaluates it for zero-short learning on image classification dataset.
MAML \cite{finn2017model} applies meta-learning (that is to train a model on multiple tasks) to  achieve good performance for few shot image classification, reinforcement learning tasks and regression tasks. 
\cite{phoo2020self} and \cite{islam2021dynamic} applies self-training based transfer learning method to use knowledge gained in one domain to solve the  few shot learning task in domains which are completely different - referred to as cross domain few shot learning task.

In the field of HAR, few shot learning has been explored in  self-supervised learning and self-training settings.
In \cite{haresamudram2021contrastive} and \cite{haresamudram2022assessing}, models pre-trained through self-supervised learning methods such as contrastive predictive coding, SimCLR and multi-task learning  are tasked to recognize activities using supervision from a limited number of labeled windows per target activity class. 
\cite{haresamudram2022assessing} compared the performance of various self-supervised, supervised and unsupervised baselines and found that SimCLR is the best performing self-supervised baseline for target datasets with varied sensor positions and target activities.
Similarly, \cite{tang2021selfhar} evaluates the self-training based semi-supervised method with a very limited labeled data availability and achieves  performance better than end to end supervised learning and self-supervised methods.

\section{Methodology: Few Shot Transfer Learning for Human Activity
Recognition}
\label{sec:method}
\begin{figure*}[t]
    \centering
    \includegraphics[width=0.9\linewidth]{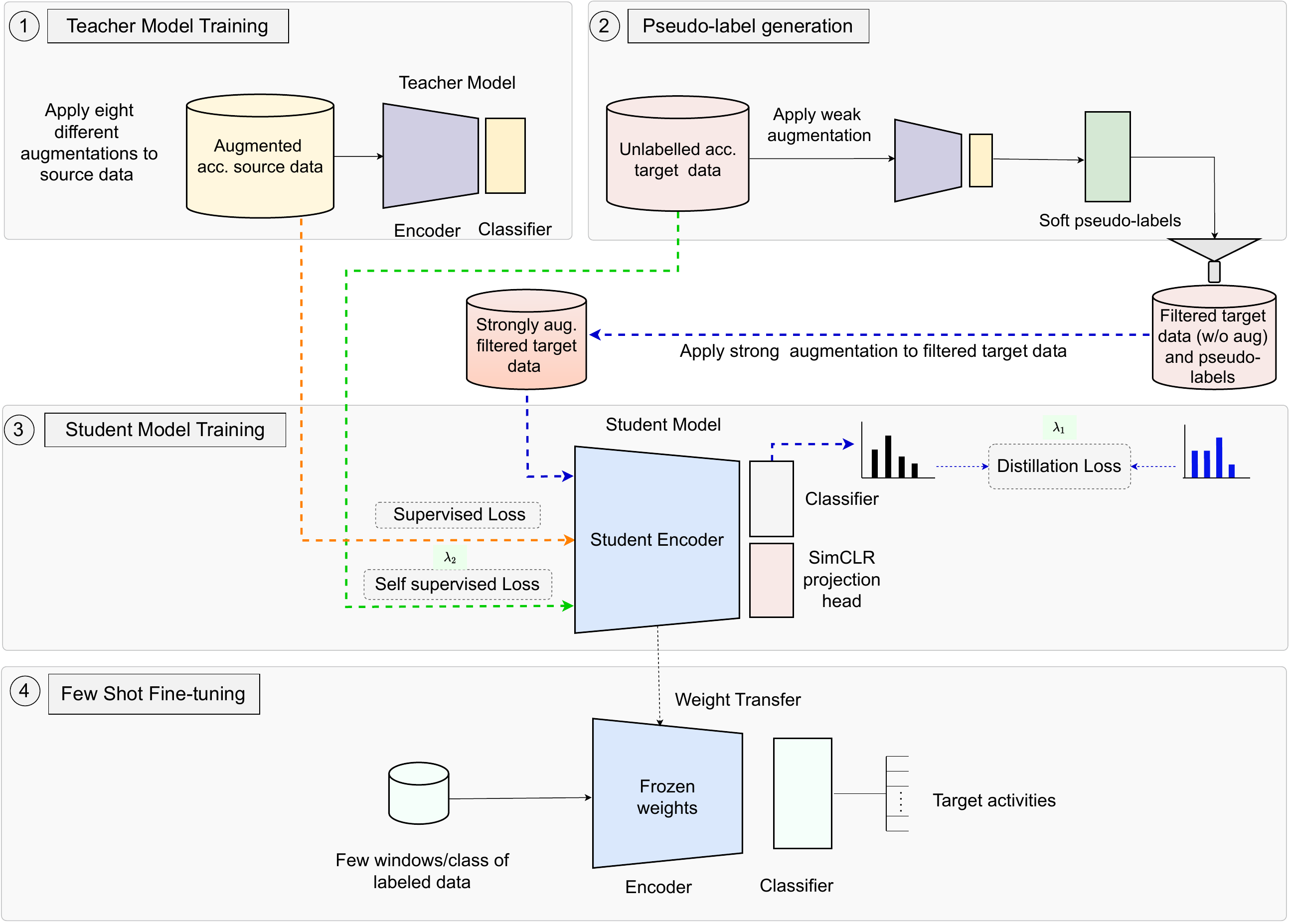}
    \caption{
    Overview of Cross-Domain HAR: It uses self-training to adapt representations learned in a labeled source domain (defined by activities, sensor positions, and users) to the target domain with very limited labeled data. 
    It consists of four stages: teacher model training, pseudo-label generation, student model training,  and few shot learning. Teacher  training involves supervised learning on the source domain data using cross-entropy loss. In the pseudo-label generation stage, the trained teacher is utilized to generate soft pseudo-labels for weakly-augmented target data. Labeled source data and unlabeled target data -- which are selected based on teacher model confidence of pseudo-label prediction--are used in further student training. Cross-entropy loss on source data, KL-Divergence loss between pseudo-labels and student model predictions of strongly-augmented target data and SimCLR self-supervised loss on target data  are jointly optimized to learn parameters of the student model. During the few shot learning stage, weights of the student model encoder are frozen and a limited number of windows per target activity class is used to fine-tune the final classifier.
    }
    \label{fig:method}
\end{figure*}

In this section, we introduce our transfer learning method -- `Cross-Domain HAR'. 
It involves teacher-student training, which uses both labeled source dataset and unlabeled target datasets to achieve effective recognition on target data.
In what follows, we formally describe the task, the technical details, and the various stages involved.
The process is summarized in Fig. \ref{fig:method}.

\subsection{Problem Setup} 
We assume the availability of a labeled source data set, $D_s$ where $D_s = \{x_0, x_1.....x_m-1\}$ contains $m$ equally-sized windows of time-series data and each window is labeled with an activity from a set of $k$ different activities, $y_s$. 
The time-series data are collected using an inertial sensor located on body position $S_1$. 
Our formulation is similar to the setup introduced in \cite{lara2012survey} and used in \cite{tang2021selfhar}. 
The goal is to adapt representations learned  from the source to novel application domains where the access to labeled data is limited.
Unlabeled target data $D_t = \{x_0, x_1.....x_n-1\}$ is collected using an on-body sensor positioned at $S_2$, which is different from source sensor placement.  
Similar to the source data, the (mostly) unlabeled target data $D_t$ consists of windows of sensor data, however with no ground truth annotations. 
In addition to varied sensor positions, the target domain can also cover a different set of activities $y_t$, which may or may not be a subset of $y_s$.
A small number of windows per target activity is labeled from the target dataset, which includes $n \in \{2,5,10,50,100\}$ windows/class. 
These limited labeled data provide supervision for fine-tuning the final recognition model to the target domain. 
The performance of the recognition system is evaluated using a target test set kept separate from the original unlabeled data $D_t$.

\subsection{Technical Details} 
Our method comprises four stages in total: 
\emph{(i)} Teacher model training;  \emph{(ii)} Pseudo-label generation;  \emph{(i)} Student model training; and \emph{(iv)} Few shot Fine-uning.
The full workflow is visualized in Fig.\ \ref{fig:method}, and in what follows, we detail each stage and its importance to improved sensor-based human activity recognition scenarios.

\subsubsection{Overall Pipeline}
The overall workflow of Cross-Domain HAR, can be summarized as follows:

\begin{enumerate}
    \item As in a typical knowledge distillation \cite{hinton2015distilling} setup, a teacher model $\theta_{t} $ is trained on a source dataset $D_s$ (top left in Fig.\ \ref{fig:method}). 
    First, source data is augmented by applying eight signal transformations (details below) to each input data window.
    Subsequently, the Teacher model is trained on this augmented labeled dataset in a supervised  manner by minimising the cross-entropy loss.
    
    \item Using the pre-trained Teacher model $ \theta_{t} $,  soft pseudo-labels are generated for weakly augmented target data points $x^{w}_{i}$ (upper right in Fig.\ \ref{fig:method}).
    Then the noisy pseudo-labels are filtered by only selecting those target data samples for which the trained teacher model makes confident predictions. 
    Pseudo-labels for weakly augmented target dataset are obtained through:
    \begin{flalign}
     y^\prime_{i} = f_{\theta_{t}}(x^{w}_{i}) \;  , \; \; \;  \forall x_{i} \in D_{t} 
    \end{flalign}
    Function $f = h \circ g  $ where g is the encoder which embeds raw data to representational space and h represents the classifier head for activity recognition.

    \item Using the labeled source dataset $D_s$, unlabeled target data $D_t$ and pseudolabels $y^{\prime}_{i}$ generated in the previous step, a new student model $\theta_{s}$ is learned by minimizing the following loss function (center part of  
    Fig.\ \ref{fig:method}):   
    \begin{align}
     L = \frac{1}{N_s}\sum{l_{CE} (x_{i}, y_{i})}_{i=0}^{N_s} + \lambda_1 \frac{1}{N_t}\sum{l_{KL} (f_{\theta_{s}}(x^{s}_{j}), y^{\prime}_{j})}_{j=0}^{N_t} + \lambda_2 l_{SIM}(D_{t})
    \end{align}

    \noindent
    where $l_{CE}$ is the cross-entropy loss on the source dataset $D_{s}$; $l_{KL}$ is the KL-divergence loss, which applies consistency regularization to match predictions for strongly-augmented target data $x^{s}_{j}$  and  $y^{\prime}_{j}$ pseudo-labels for weakly augmented data; and $l_{SIM}$ is the SimCLR loss on the target data $D_{t}$.
    The loss formulation is similar to \cite{phoo2020self}. 
    Here, $\lambda_1$ and $\lambda_2$ are used to balance the KL-divergence loss and SimCLR losses, and values are determined via hyper-parameter tuning.
    
    \item  The learned weights of the Student model's encoder are frozen and few labeled windows per class from target dataset are used to fine-tune a new classifier (bottom part of Fig.\ \ref{fig:method}).
\end{enumerate}

\subsubsection{Importance Of Each Loss Term:}

The student model is trained by jointly optimising over three different loss terms:
\emph{(i)} Cross-entropy loss on the source dataset $D_{s}$: which assists the student model in learning the classification task as learned by the Teacher model itself, and comprises the standard task loss learned in self-training setups \cite{xie2020self};
\emph{(ii)} KL-divergence loss: which is tasked to match the Student model predictions for strongly augmented unlabeled data to teacher model predictions for weakly augmented unlabeled data.
This invariancy constraint on Student model predictions ensures that the Student model learns more than the Teacher, as it is forced to learn activity classes for difficult signals \cite{xie2020self}.
Additionally, adding different augmentations/perturbations to the input imposes input-consistency regularization on the learner.
Previous work  \cite{wei2020theoretical} proved theoretically that the use of consistency regularization in the self-training setup achieves higher accuracy.
It was also shown that  when using self-trained models,   which are  consistent w.r.t. input transformations, the  correctly pseudolabeled examples  will denoise examples with incorrect pseudolabels and thus result in robust performance; and
\emph{(iii)} SimCLR-based self-supervised loss on the target dataset $D_{t}$: similar to \cite{phoo2020self}, our method needs to bridge a considerable domain gap introduced by differing on-body sensor positions and downstream activities, which is challenging. 
Combining the self-training with self-supervision on unlabeled target data aids in drawing out additional information from target domain.
Therefore, we leverage both existing labeled data, as well as more easy-to-collect unlabeled data for building activity classifiers with low cost and annotation effort.

\subsubsection{Teacher Model Training} \label{teacher_training}
The first step involves training the Teacher model to recognize activities, using available, annotated data.
This involves training a multi-class activity recognition model for the activities present in the source dataset. 
It is essential that the Teacher model learns generalized features that can be successfully adapted to diverse target domains, even across differing data distributions.

To extract generalized sensor representations, it would be ideal to utilize a large-scale source dataset based on various on-body sensor placements and orientations, diverse activities and users, as well as heterogeneous devices.
Yet, most publicly available, labeled datasets in sensor-based HAR are rather limited, for example, containing only a few users and activities \cite{kwon2020imutube}.
In order to  inject diversity and variation into the source data, we employ eight data transformations designed for accelerometry in \cite{um2017data}:
\emph{(i)} adding random Gaussian noise; 
\emph{(ii)} scaling signal by a random factor; 
\emph{(iii)} applying a random 3D rotation; 
\emph{(iv)} reversing the time direction of input signal; 
\emph{(v)} negating input signal values; 
\emph{(vi)} warping the signal; 
\emph{(vii)} shuffling channels randomly; 
\emph{(viii)} randomly perturbing parts of time-series signals.\footnote{Code for transformations used from https://github.com/iantangc/ContrastiveLearningHAR  }
After augmentation, the source data is increased in size by a factor of 9 (due to the eight transformations applied in addition to the original data), and is used for end-to-end training.

\subsubsection{Pseudo-label Generation} 
\label{pseudolabel_generation}

The trained Teacher is used to predict soft pseudo-labels (i.e., a probability distribution over activity classes) on  the target data. 
The unlabeled target windows are first weakly-augmented before inferring the output probabilities from the Teacher model.
For generating weakly augmented views,  we apply random 3D rotation to the unlabeled target data. 
The weak augmentation is performed to add consistency regularization to the self-training process (more details provided in Sec. \ref{student_training}). 
As hypothesized in \cite{phoo2020self}, although the source and target activities and locations may be different, the pseudo-labeling process generates groupings on the target unlabeled data, thereby improving the few shot performance.
Subsequently, we filter the pseudo-labeled target data based on Teacher model confidence.

As the pseudo-labels generated by Teacher model are used by the Student model as target predictions, it is advantageous to select only the confidently predicted samples to avoid learning from noisy labels. 
Therefore, we filter windows with prediction probability $<30\%$ before performing student model training. 
Given that, for example, the source dataset Mobiact has $11$ activities, we found that a maximum probability threshold of 30\% is practical and higher than making random predictions (i.e, threshold of $\approx9\%$).
A more stringent threshold value may lead to reduced data sizes, but it may not be  sufficient for Student model training.

\subsubsection{Student Model Training}
\label{student_training}
The features learned by the Teacher model are specific to the source domain and its activities. 
To adapt these representations to target data--across a substantial domain gap--the Student model is trained by minimizing a sum of:  
\emph{(i)} Cross-entropy;
\emph{(ii)} KL-divergence; and 
\emph{(iii)} the Self-supervised losses. 

The  cross-entropy loss is calculated on labeled augmented source data, which is similar to how the Teacher model is trained to learn source feature representations. 
The KL divergence loss reduces the distance between Teacher model predictions and Student model output,  and  acts as the distillation loss, ensuring that the knowledge from teacher model is distilled/transferred to student network. 

We add consistency regularization to our self-training process, in which the Teacher model generates  pseudo-labels for weakly-augmented versions of target data, whereas the Student model is tasked to predict activities for strongly-augmented versions of the same data.  
For strong augmentation, we generate a heavily distorted view of the input sensor data by applying multiple transformation functions one after another on the input signal, including: 
\emph{(i)} adding random Gaussian noise; 
\emph{(ii)} applying random 3D rotation; and 
\emph{(iii)} negating the input sensory values. 
Adding random rotation, noise, and negation to sensor data can be useful in learning representations that are invariant to sensor heterogeneity, orientation/placement, and offset variances (can potentially capture some of the data recording differences across domains)\cite{saeed2019multi}. 
Predicting for heavily distorted views of data ensures that the Student model learns beyond the Teacher model, and can adapt more easily to target domains (whose data may not be similar to source).

In addition to the self-training loss and distillation losses \cite{xie2020self}, we also add a self-supervised loss on the target dataset -- similar to the protocol followed in \cite{phoo2020self}. 
The self-supervised loss on unlabeled target dataset is utilized to aid the Student model in learning target domain specific useful features. 
We use state-of-the-art contrastive self-supervised learning, SimCLR \cite{chen2020simple, tang2020exploring}, which applies two augmentations to obtain correlated views of the same signal data, and maximizes the similarity between the representations of these views using NT-Xent loss (normalized temperature-scaled cross entropy).

\subsubsection{Few Shot Fine-tuning}
Representations learned by the Student for the target domain are further fine-tuned to the downstream recognition task. 
We freeze the encoder weights of the Student model and attach a new classifier, which is trained only using a small number of randomly drawn labeled windows per activity class. 
We perform 5-fold cross validation across 5 randomized runs and report mean/standard deviation of macro F1-scores.

\vspace{2mm}
In Cross-Domain HAR, we leverage existing labeled datasets to first learn useful representations of the source domain.
The goal of our method is to adapt the representations learned from the source domain to target conditions, which may be substantially different from the source, i.e., to learn \emph{generalizable} features.
Further, the target data are mostly unlabeled, apart from few seconds/minutes with annotations, e.g., performed  through user input without significant disruption of user experience.
Under these conditions, we make maximal use of the available unlabeled data by not only performing traditional self-training, but also by adding self-supervision to add as much target domain specific information as possible to the features, resulting in more effective representations for the target domain.
Therefore, our method results in more accurate recognition of human activities, without the reliance of costly annotations for the target domain, which can be of high practical utility wearables applications.

\section{Setup}

In previous sections, we motivated our work on developing semi-supervised methods that can leverage  existing  labeled HAR datasets to bootstrap recognition models for new application domains with unseen activities and on-body sensor locations. 
In this section, we provide the details of our method settings and implementation, which includes: 
\emph{(i)} Description of datasets used for evaluation; 
\emph{(ii)} Data pre-processing;
\emph{(iii)} Model architecture;
\emph{(iv)} Hyperparameters;
and \emph{(v)} Evaluation metric.

\subsection{Datasets}

\begin{table}[t]
	\centering
	\caption{
		Summary of the datasets used for our experimental evaluation:
        Pre-training is performed using Mobiact  (unless specified differently), while the remaining datasets are used as targets.
        In order to study the performance across sensor positions, we chose two datasets per target location: wrist, leg/ankle, and waist/trousers, covering diverse activities and dataset sizes.
        This table has been adopted/adapted with permission from \cite{haresamudram2022assessing}.
	}
	\small
    \resizebox{\textwidth}{!}{%
	\begin{tabular}{P{2.2cm} P{1.5cm} c c P{8.0cm}}
        \toprule 
		Dataset & Location & \# Users & \# Act. & Activities \\ 
		\midrule 

        \textbf{Mobiact} \cite{chatzaki2016human} & Waist/ Trousers & 61 & 11 & Standing, walking, jogging, jumping, stairs up, stairs down, stand to sit, sitting on a chair, sit to stand, car step-in, and car step-out\\ 
        \hline
  
		\textbf{HHAR} \cite{stisen2015smart} & Wrist & 9 & 6 & Biking, sitting, going up and down the stairs, standing, and walking \\ 
		%
		\textbf{Myogym} \cite{koskimaki2017myogym} & Wrist & 10 & 31 & Seated cable rows, one-arm dumbbell row, wide-grip pulldown behind the neck, bent over barbell row, reverse grip bent-over row, wide-grip front pulldown, bench press, incline dumbbell flyes, incline dumbbell press and flyes, pushups, leverage chest press , close-grip barbell bench press, bar skullcrusher, triceps pushdown, bench dip, overhead triceps extension, tricep dumbbell kickback, spider curl, dumbbell alternate bicep curl, incline hammer curl, concentration curl, cable curl, hammer curl, upright barbell row, side lateral raise, front dumbbell raise, seated dumbbell shoulder press, car drivers, lying rear delt raise, null \\ 
		\midrule

        \textbf{RealWorld} \cite{sztyler2016body} & Waist & 15 & 8 &  Sitting, standing, lying, walking, running, stairs up, stairs down, jumping \\ 
		%
		\textbf{Motionsense} \cite{malekzadeh2018protecting} & Waist/ Trousers & 24 & 6 & Walking, jogging, going up and down the stairs, sitting and standing \\ 
		\midrule
  
		\textbf{MHEALTH }\cite{banos2014mhealthdroid} & Leg/ Ankle & 10 & 13 & Standing, sitting, lying down, walking, climbing up the stairs, waist bend forward, frontal elevation of arms, knees bending, cycling, jogging, running, jump front and back\\ 
  %
		\textbf{PAMAP2} \cite{reiss2012introducing} & Leg/ Ankle & 9 & 12 & Lying, sitting, standing, walking, running, cycling, nordic walking, ascending and descending stairs, vaccuum cleaning, ironing, rope jumping \\ 
		
        \bottomrule
	\end{tabular}
 }
	\label{tab:datasets}
\end{table}

We work with seven datasets, where Mobiact is the labeled source (unless specified differently), and the rest serve as targets.
In what follows, we briefly describe the datasets (summarized in Tab.\ \ref{tab:datasets}).

\subsubsection{Source Dataset}
We use the Mobiact v2 dataset \cite{chatzaki2016human} as the primary source for our experiments (unless specified differently).
It was recorded using a smartphone's inertial measurement unit while participants performed different activities.
The phone was placed freely in the pocket of the participant's choice and in random orientation, in order to simulate everyday use.
The dataset contains four types of falls and 12 activities of daily living, from a total of 66 subjects.
Of those, 51 were men and 15 were women with ages between 20-47 years, height ranging in 160-193 cm, and the weight between 50-120 kg.
As in \cite{saeed2019multi}, we include 11 activities in our analysis (i.e., exclude lying, see Tab.\ \ref{tab:datasets} for details), and also do not analyze the falls, resulting in $61$ subjects in total.
 Mobiact is chosen as the source due to its diverse nature w.r.t.\ participants as well as the collected activities, thereby aiding in the development of generalizable features.
 
\subsubsection{Target Datasets} 
We evaluate the performance on six target datasets, covering a range of activities and sensor positions. 
For some targets, the covered activities do not overlap with those from the source, whereas there is overlap for other scenarios. 
Similarly, we also consider on-body sensor locations which are not the same as the source, i.e., not at the waist.
These datasets are chosen such that we can study and demonstrate how our method bridges varying degrees of domain gaps between source and target.

We only utilize data from a single accelerometer placed at specific positions, even if some datasets contain multiple sensor modalities and locations. 
The data are down-sampled to 50 Hz to match with the lowest frequency among all datasets.
The description of the target datasets is provided below, grouped by sensor locations:

\paragraph{Wrist/Arm:}
    \begin{enumerate}
        \item HHAR \cite{stisen2015smart}: 
        This dataset covers six locomotion-style activities collected via smartwatches by nine users aged 25-30. 
        Most of the activities covered are simple everyday activities as they are  prominently studied for many HAR application scenarios. 
        There is substantial overlap in activities with Mobiact, although biking and sitting are not covered.
        
        \item Myogym \cite{koskimaki2017myogym}: 
        Thirty fine-grained gym exercises performed by ten users were recorded using a Myo Armband. 
        The resulting dataset is highly imbalanced with 77\% of data categorized as the $NULL$ class. 
        There is no overlap of activities between our source Mobiact and Myogym. 
    \end{enumerate}
    
\paragraph{Waist/Trousers:}
    \begin{enumerate}
         \item RealWorld \cite{banos2014mhealthdroid}: The RealWorld dataset was collected to cover common everyday activities such as sitting, standing, and walking, some of which are  covered by  source dataset.
         Fifteen subjects including 8 males and 7 females participated in the data collection process, with ages $20-44$, height and weight around $173.1 \pm 6.9$cm and $74.1 \pm 13.8$kg,  respectively.
         The acceleration data is collected from 7 different body positions, however we only consider the data collected at the waist.
         
         \item MotionSense \cite{malekzadeh2018protecting}: 
         The accelerometer, gyroscope and altitude data were collected using an iPhone 6s kept in the trouser's front pocket of each participant.
         A total of 24 participants recorded the data. 
         Weight range:  $48$-$108$ kg;
         Height:  $161$-$190$cm;
         $10$ females and $14$ males;
         Age: $18-46$ years. 
         Six locomotion-style activities including walking and running were recorded, and are fully contained in the source dataset.

    \end{enumerate}
    
\paragraph{Leg/Ankle:}
    \begin{enumerate}
         \item MHEALTH \cite{banos2014mhealthdroid}: 
         This dataset covers locomotion-style activities such as standing, sitting, walking, in addition to exercises like jumping front and back, bending knees, etc. 
         The aim of the study was to build a novel framework for mobile health applications and ten users participated in the study.
         Similar to Myogym (see above), the dataset is imbalanced with a large null class.
         Some of the target activities such as waist bend forward, knees bending and frontal elevation of arms are not seen in the source datasets.
         
         \item PAMAP2 \cite{reiss2012introducing}: 
         This dataset covers twelve activities of daily living, including domestic activities (such as vacuum cleaning, ironing) and exercises (nordic walking, cycling). 
         Some of these activities are covered in our source while many others (such as ironing and Nordic walking) are not.
         The dataset has over ten hours of recording, from nine users, of which eight were male and one was female. 
    \end{enumerate}

\subsection{Implementation Details}
Here, we discuss the practical details of our method, including the data pre-processing, model architectures, hyperparameters used in our experiments, and the evaluation metric used to measure performance.

    \subsubsection{Data Preprocessing:}
    We use raw  accelerometer data for each dataset collected at specific positions, as we only study single sensor scenarios.
    The data are split by participants such that 20\% of the users are randomly chosen to form the test set, whereas the remaining users are once again partitioned randomly at a 80:20 ratio to comprise the train and validation splits.
    Using this protocol, we create five separate folds, such that each user in the dataset appears only once in the test set overall (across folds).
    The mean of the test set performance across folds is reported in subsequent sections. 
    
    We only apply minimal data pre-processing to the sensory data, and downsample them to 50 Hz in order to match the lowest frequency across datasets.
    Further, we apply sliding window based segmentation with a window size of 1 second and overlap of 50\%.
    Therefore, each window has a size of 50 times 3, which includes data from 50 timesteps and 3 accelerometer channels. 
    
    In the source dataset, the train data are normalized to have zero mean and unit variance. 
    Subsequently, the means and variances are used to normalize the validation and test splits of source data as well.
    The five different folds of target dataset are also normalized using the source data statistics  instead of using target data means and variances as this leads to performance improvement \cite{haresamudram2022assessing}.
    
\subsubsection{ Model Architecture:} 
\label{model_arch}
The three training stages--Teacher model training, Student model training, and Few shot learning--use a common encoder, which learns representations from data.
Depending upon the training strategy of each stage, different task-specific heads are attached to the encoder model. 
It consists of three convolutional blocks with (32, 64, 128) filters, and a kernel size of (24, 16, 8) respectively, followed by global maximum pooling.
ReLU activation function \cite{nair2010rectified} and Dropout \cite{srivastava2014dropout} of p=0.1 is applied to output of each convolutional layer.

During the Teacher model training, we attach a Multi-Layer Perceptron (MLP) to the output of the global maximum pooling layer of encoder model. 
The MLP consist of three linear layers with ReLU activation, Batch Normalization \cite{ioffe2015batch}, and Dropout (p=0.2) applied between each of these layers.
Once the Teacher training is completed, the learned encoder weights are transferred to the Student model, which has two task-specific heads.
First, a Classifier head (same as in the Teacher model) is used to outputs logits for source dataset as well as the strongly-augmented target dataset.
Second, the projection head for SimCLR which consists of a 3-layer feedforward network with the first two layers comprising 256 and 128 units respectively, and the ReLU activation after each of the first two layers.
The last layer generates a 50 dimensional representation for each data point which is used for calculating the self-supervised loss. 
The weights learned by the Student model are frozen and a randomly initialized classifier head is fine-tuned with few labeled windows/class of target data. 
The classifier head for evaluation with limited labeled data is an MLP with three layers, identical to the Teacher model, albeit the number of units in the last layer equals the number of classes in the target dataset.

We contrast our method's performance against Contrastive Predictive Coding (CPC) \cite{oord2018representation}, which functions as a self-supervised baseline.
The architecture is identical to \cite{haresamudram2021contrastive}, and contains an encoder with three convolutional blocks with (32, 64, 128) filters and same kernel size of 3. 
Each convolutional block is followed by ReLU activation and Dropout with p = 0.2.
The encoder is followed by a two layer GRU which is used as the auto-regressive model to get the context vector, along with a number of  linear prediction networks to predict the future timesteps.

We also compare the performance against SimCLR \cite{chen2020simple, tang2020exploring}, and use the transformations as deployed by \cite{tang2020exploring} for the contrastive learning HAR setup.
The encoder is identical to the one used in our method, and is composed of convolutional blocks with (32, 64, 128) filters, and a kernel size of (24, 16, 8), respectively.
The output of the last convolutional block is passed through a  global maximum pooling layer before the projection head projects the embedding to a 50-dimensional space.
The projection head consists of three linear layers with 256, 128 and 50 output units respectively, and with ReLU activation function applied between the linear layers.

\subsubsection{Hyperparameters:}
We use the PyTorch \cite{paszke2019pytorch} framework for all implementations.
Our workflow consists of three training stages, i.e., Teacher model, Student model and Few-shot training, where the performance of each stage depends on the preceding stage(s). 
For the supervised training of the Teacher model, we used a learning rate of $1e-3$ and L2 regularization $1e-5$ for Mobiact and learning rate of $5e-4$ and L2 regularization of $1e-5$ for HHAR, as detailed in \cite{haresamudram2022assessing}.

To find suitable loss coefficients for Student model training, we perform a random search of 50 parameter combinations over a range of values for the learning rate, L2 reg.\ , batch size, $\lambda_1$ and $\lambda_2$ (refer Tab. \ref{tab:hyperparam}).
Subsequently, we perform Student model training with these combinations.

Further, for each of the  Student model parameter combinations, we randomly search 15 hyperparameter combinations for Few-shot fine-tuning (i.e., the learning rate and L2 regularization).
The best overall combination of hyperparameters across Student model training and the Few-shot fine-tuning is obtained using the validation F1-score over one of the target dataset folds when the number of randomly sampled windows / class is 25. 

We find that $\lambda_1=0.7$ and $\lambda_2=0.8$ performs best. 
In order to simplify the parameter search and to demonstrate the robustness of our method across source and target datasets, we  perform $\lambda_1$ and $\lambda_2$  search only  for PAMAP2 as the target dataset and HHAR as the source.
For all other experiments (across different source and target datasets), the loss coefficients obtained above are fixed before tuning for the rest of hyper-parameters for Student model training.

For the pseudo-label selection, we performed a preliminary analysis over different confidence threshold values, and found 30\% to be reasonable for filtering out less confident samples. 

The hyperparameter combinations are detailed in Tab.\ \ref{tab:hyperparam} for reference and briefly described below.
For Contrastive Predictive Coding (CPC) based pre-training using Mobiact, we used a learning rate of $5e-4$, number of steps for prediction $28$  and batch size of $64$, as in the original paper \cite{haresamudram2021contrastive}. 
When using HHAR as the source, we performed a random search (10 random samples) over a range of batch sizes, number of future timesteps, learning rates and L2 regularization. 
In the case of  SimCLR, we performed grid search (as total combinations are not very large) with larger batch sizes. 
Similarly, for self-supervision, naive transfer and random initialization methods, the Few-shot fine-tuning performance is evaluated by conducting a grid search over the hyperparameters.

We used a batch size of 256  for the Teacher model training. 
For Few-shot fine-tuning, we create batches in accordance with total  data size according to number of labeled windows available ( say to fine-tune the model with two labeled windows per activity class for HHAR target dataset with six activity classes, we chose batch size as four to ensure two to three passes of different batches).

\subsubsection{Performance Metric}

The mean test set F1-score (i.e., macro F1-score) is utilized as the performance metric due to its resistance to class imbalance (which is common in wearables-based HAR datasets) \cite{kwon2020imutube, powers2020evaluation, plotz2021applying}: 

\begin{equation}
	F = \frac{2}{|c|}\sum_{c}^{} \frac{prec_{c} \times recall_{c}}{prec_{c} + recall_{c}}
\end{equation}
where $|c|$ is the number of classes, and $prec_c$ and $recall_c$ are the precision and recall for each class respectively.

\section{Results and Discussion}
\label{sec:results}

\subsection{Cross-Domain Human Activity Recognition}
We fine-tune the representations derived through our teacher-student learning method using few labeled windows per target activity class.  
These features are subsequently evaluated on a test set drawn from target domain, which is kept separate from the training process. 
We perform five-fold cross validation for five random runs in this few shot evaluation.
For each random run, the model is trained using randomly sampled,  annotated windows per activity from the train split of the target dataset fold, and the mean F1-score is calculated on the (untouched) test set. 
This is repeated for the five folds, and the test set F1-score across folds is first averaged and then used to report the mean and confidence interval, i.e., standard deviations, (across random seeds) in Fig.\ \ref{fig:results}.

We demonstrate the performance of our `Cross-Domain HAR' against the following baseline methods:
\begin{enumerate}
    \item \emph{Contrastive Predictive Coding} \cite{oord2018representation, haresamudram2021contrastive}, denoted as CPC in Fig.\  \ref{fig:results}, where we pre-train the encoder in a self-supervised manner using the CPC protocol \cite{haresamudram2021contrastive} for time-series signal data.
    The architecture of encoder, autoregressive network, and prediction networks is described in Sec.\ \ref{model_arch}, and is identical to \cite{haresamudram2021contrastive}.
    For pre-training, we use augmented source data as described in Sec.\ \ref{teacher_training}, similar to Cross-Domain HAR.  
    
    \item \emph{SimCLR}  \cite{chen2020simple}, through which we learn features by maximizing the similarity between augmented versions of the input.
    The architecture of the encoder and projection head is provided in Sec.\ \ref{model_arch}, and  similar to \cite{tang2020exploring}. 
    For SimCLR we did not use the augmented source data as the pretext task already involves applying random transformations to windows of sensor data. 
    The same set of eight transformations are applied for SimCLR as well (as detailed in Sec.\  \ref{teacher_training}).
    
    \item \emph{Naive transfer}, which comprises evaluating the features learned by the encoder from labeled source data. 
    First, a convolutional encoder and an MLP classifier identical to the Teacher model (see Sec.\ \ref{model_arch}) are trained end-to-end using labeled source data. 
    Subsequently, the encoder weights are frozen and  and used for predicting activities on the target data. 
    Once again, we train the encoder with augmented source data.
    
    \item \emph{Supervised learning}, where we use a convolutional encoder and an MLP classifier head (referred to as `Conv.\ Classifier' in Fig.\  \ref{fig:results}), directly trained on target labeled data for HAR.
    The architecture is identical to the Teacher model, albeit the size of the last layer equals the number of target dataset classes.
    
\end{enumerate}

\subsubsection{Using Mobiact as source dataset:}
\label{sec:smobi_results}

\begin{figure*}[h!]
    \begin{subfigure}[t]{0.8\textwidth}
        \includegraphics[width=0.99\linewidth,trim = {0, 0, 0, 0.2cm}, clip]{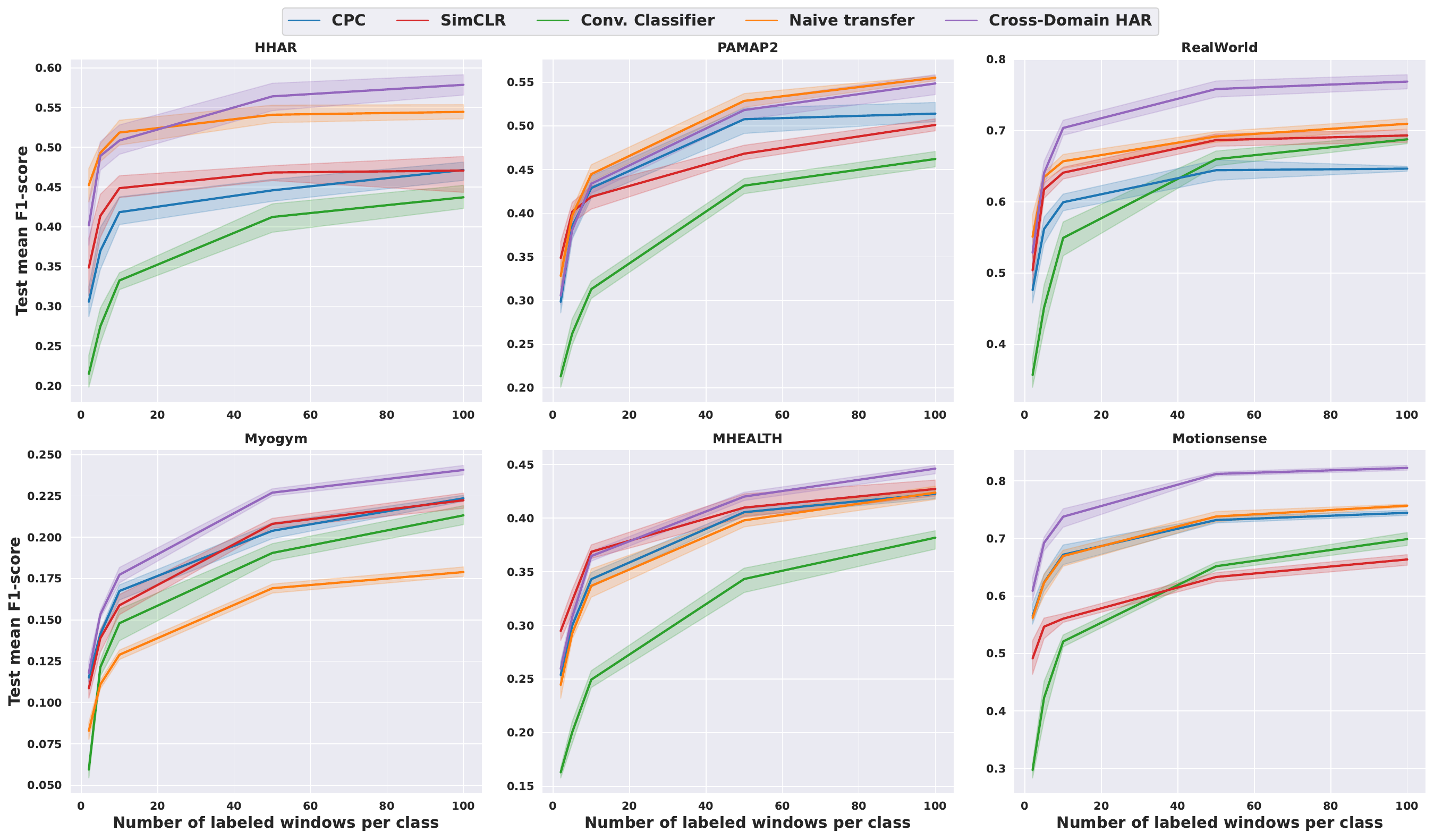}
        \caption{\label{fig:smobi_results}Mobiact as source.}
    \end{subfigure}
    \begin{subfigure}[t]{0.8\textwidth}
        \includegraphics[width=0.99\linewidth,trim = {0, 0, 0, 1.2cm}, clip]{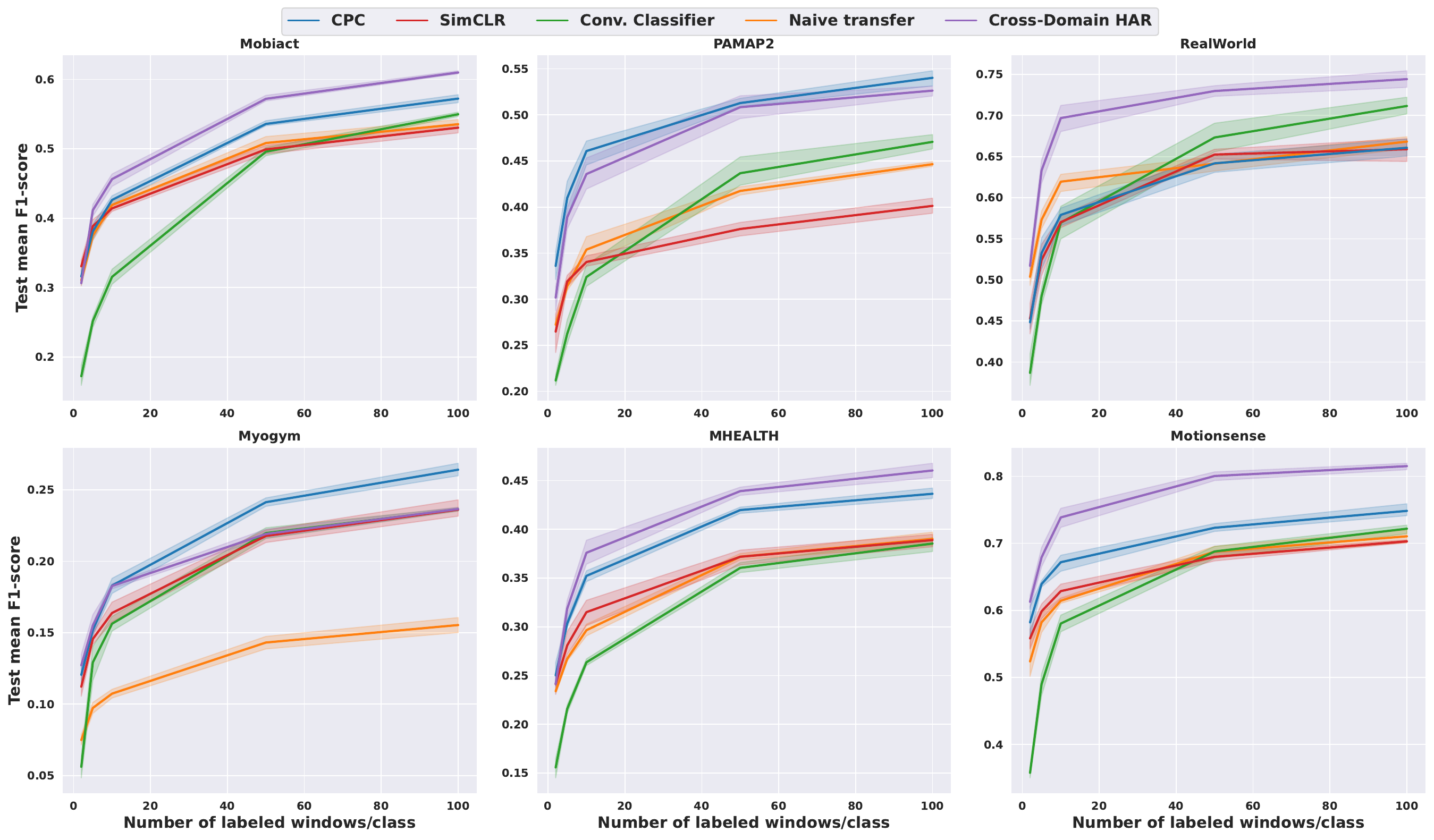}
        \caption{\label{fig:shhar_results}HHAR as source.}
    \end{subfigure}
    \vspace*{-1em}
    \caption{
    HAR performance for  target datasets with limited annotations. 
    For CPC and SimCLR, the encoder is pre-trained (self-supervised) using source data. 
    Encoder weights are frozen and a randomly initialized classifier is fine-tuned with a limited  number of labeled windows n $\in \{2,5,10,50,100\}$.
    The same process 
    is followed in Cross-Domain HAR, however the encoder weights are learned using our self-training approach. 
    In the Naive transfer approach, the encoder is pre-trained (supervised) on the source domain.
    Conv.\ Classifier is trained end-to-end from scratch using the same number of windows per activity.
    Cross-Domain HAR performs significantly better than  other approaches for most target datasets, across varied sensor positions and activities. 
    }
    \label{fig:results}
\end{figure*}

For each of the baselines, except  Conv.\ Classifier, the features learned by the encoder are frozen and used for classification on the target domains with an MLP network, whereas for the Conv.\ classifier the network is trained end-to-end.
Results of our evaluation are shows in Fig.\ \ref{fig:smobi_results}.
We first compare the results of our method using Mobiact as the source dataset, which was collected at the waist and contains mostly locomotion-style activities and some transitional classes.
We begin with Mobiact because it is the largest dataset under evaluation, comprising 61 participants in total. 

First, we compare the performance of Cross-Domain HAR against the self-supervised baselines, which are CPC and SimCLR.
For most target datasets, the proposed method outperforms the self-supervised methods by substantial margins.
For the wrist-based HHAR, we obtain  a performance increase of around 6-7\% and 10-11\%, respectively, when 2 to 5 and 50 to 100 windows of annotated activity data are available per class.
For Myogym, where we try to cover a large domain gap across activities (as Myogym has fine-grained gym activities, which are  disjoint from locomotion-style activities in Mobiact) and across sensor locations (waist for Mobiact vs wrist for Myogym), our method achieves a  improvement of around 2\% across all configurations of labeled windows/class.
In the case of RealWorld and Motionsense, which cover the same sensor position but different activities  when compared to the source (Mobiact), we achieve significant improvements over self-supervised methods by 10-12\% and 7-8\% respectively.

When the number of labeled windows/class is very small, i.e., $n \in \{2,5,10\}$, self-supervised baselines perform slightly better for ankle-based MHEALTH, whereas our method performs better when there is access to more annotated data. 
For the other ankle-based dataset--PAMAP2--we observe a similar trend where our method outperforms SIMCLR ($\approx{5\%}$) as well as CPC ($\approx{2-3\%}$) when 50 to 100 windows of annotated activity data are available per class.

Interestingly, using Mobiact as source dataset, we found Naive transfer to be the best performing baseline for four of the six target datasets. 
Therefore, simply training end-to-end on the Mobaict dataset followed by fine-tuning on the target is sufficient for high performance.
However, for most target datasets except PAMAP2, Cross-Domain HAR outperforms Naive transfer, with increases of approx.\ 7\% for Motionsense, 7-8\% for RealWorld and 2-3\% for MHEALTH and HHAR. 
For PAMAP2, Cross-Domain HAR lags by around 1\% when compared to Naive Transfer.

We note that for both MHEALTH and Myogym, Naive Transfer is one of the least effective baselines.
We posit that as both Myogym and MHEALTH have a large \textit{NULL} class, it is challenging to predict them in a transfer learning setup as activity specific features were learned on a source dataset with no \textit{NULL} class.
Additionally, the movements in \textit{NULL} classes across datasets may not be similar.

We  note that our method's performance can be highlighted well by comparing it to  Naive transfer.
This is because Naive transfer essentially involves fine-tuning the learned weights from the Teacher model using an MLP classifier. 
Therefore, when Naive transfer performs poorly, the features learned from the source dataset are not suitable for a target domain, and a significant gap is likely needed to be covered to adapt these representations for reasonable performance in target settings.
On the other hand, our method is able to bridge this gap as it pushes the recognition performance over naive transfer for most the target datasets, thereby indicating the necessity of the other components such as self-training, consistency regularization and self-supervision.

When compared with the supervised baseline--Conv.\ Classifier--our method shows significant increase in performance. 
For scenarios where training is performed using just two to five windows/class of labeled activity data, we observe a  difference of around  20\% for HHAR and RealWorld, and 9-10\% for PAMAP2 and MHEALTH.
With increase in number of labeled windows/class, as expected, the performance of Conv.\ Classifier starts increasing and it surpasses or becomes comparable to other baselines incl.\ SimCLR, CPC, and Naive transfer.  
However, for none of the datasets it outperforms Cross-Domain HAR, which clearly demonstrates the quality of representations learned by our method.
Therefore, on a setting of great importance to wearables applications--limited availability of  labeled example data--we observe the superior capability of our method as it not only performs better than self-supervision but also than end-to-end training and performing supervised transfer learning.

We note that for different sets of target activities, users and sensor positions, our method  outperforms other baselines in most cases, and thereby 
demonstrates the capability of our method in leveraging both source, labeled and target, unlabeled data towards effective  recognition performance.

\subsubsection{Using HHAR as source dataset:}

To evaluate our method for a different set of source-target conditions, we provide results for using HHAR as the source dataset (Fig.\ \ref{fig:shhar_results}).
All methods, experiment settings as well as target datasets are identical to  Sec.\ \ref{sec:smobi_results}. 
Studying the performance relative to the self-supervised baselines--SimCLR and CPC--we note that Cross-Domain HAR performs better for most target datasets -- with the exception of CPC for PAMAP2 and Myogym.
For RealWorld, PAMAP2 and Motionsense, our method's performance is higher than SimCLR by approx.\ 10-12\%, whereas for the ankle-based MHEALTH, the performance increases by around 5-6\% with the availability of larger quantities of annotations (>10 labeled windows per class).

Using  HHAR as source dataset, we observed that CPC is the best performing baseline for most of the target datasets.
Our method, Cross-Domain HAR consistently outperforms  CPC for four out of six target datasets. 
The performance increase is significant for many datasets, which is  approx.\ 10\%  and 5-6\%,  respectively for RealWorld  and Motionsense  and  around  to 2-3\% for both  MHEALTH and Mobiact.
In case of wrist based Myogym, our method performs comparably for fewer numbers of  labeled  windows per activity, however CPC surpasses our method's performance and achieves the best F1-score for more than 10 labeled windows. 
For PAMAP2, CPC is the best performing method but our method closely follows its performance and lags behind by 1-2\%  across different number of labeled windows.

When comparing end-to-end training (via the Conv. Classifier), we observe that Cross-Domain HAR achieves much higher F1-scores.
For example, the F1-score  that our method  achieves for just two labeled windows per activity increased by around 26\% for Motionsense, 13\% for Mobiact and  15\% for RealWorld.
As the number of labeled data points increases, end-to-end performance increases, however it is unable to match the  performance of our method demonstrating the effectiveness of our approach.
Relative to Naive transfer, our method  performs better for  all datasets, increasing the performance by 6-7\% for PAMAP2, Myogym and RealWorld, 9-12\% for Motionsense dataset. 
We note that naively transferring to differing target conditions results in lower  performance when using HHAR as the source (instead of Mobiact, a larger and more diverse source dataset).
The significant improvement achieved by Cross-Domain HAR over Naive transfer when HHAR is the source shows that our method is able to better cover the domain gap induced by differing source and target activities and sensor locations, especially when using a source dataset which is smaller in size.

Overall, Cross-Domain HAR using HHAR as the source performs  very well for all target datasets.
Especially for RealWorld, the performance is impressive, considering that it is bigger with 15 users and 8 activities, whereas HHAR only covers nine users and locomotion-style activities. 
Even for Mobiact, which covers a larger activity set and 61 users, and MHEALTH, with  large \textit{NULL} class, the method is effective when trained on HHAR as the source.
Given that HHAR comprises locomotion-style activities and is smaller, the  performance gains achieved for all target datasets across sensor locations and activities shows that our method can leverage different kinds of source datasets with varying activities and sizes. 
This demonstrates the generalization capability of combining self-supervised and semi-supervised methods to derive models with superior HAR performance.

\subsection{Insights Derived from Cross-Domain HAR}
In the previous section, we have evaluated the performance of Cross-Domain HAR for transferring to a  diverse set of target domains using two source datasets. In what follows, we analyze these results to give insights into differing source-target conditions which impact successful transfer. This includes analyzing: \emph{(i)} Choice of source dataset; and, \emph{(ii)} Transferring to different target sensor positions and activities.

\subsubsection{Choice of source dataset:} \quad

\noindent
When comparing the performance of Cross-Domain HAR using Mobiact or HHAR as the source dataset, we observe that the performance achieved by both source datasets is comparable.
For example, using  100 labeled windows/class, both Mobiact- and HHAR-based settings lead to a mean F1-score of around 45\%  for MHEALTH, 23\% for Myogym, 82\% for Motionsense and 54\% for PAMAP2. 
However for RealWorld, Mobiact source dataset achieves 3-4\% higher mean F1-score.
We posit that these differences can be attributed more to the different source-target conditions rather than solely to the source dataset characteristics.

HHAR and Mobiact comprise vastly different conditions -- Mobiact is a larger, diverse dataset with sixty-one participants covering eleven different activities, while HHAR is a smaller dataset with only nine users and six activities.
Using a more diverse dataset containing more users and activities presents a better option as it would capture varied human movements and other idiosyncracies, which can assist in  generalizing and transfering to a range of different target datasets.
Still, our method is able to utilize even smaller existing labeled datasets to drive performance for target domains with limited label availability – successfully achieving the objective of this work.

\subsubsection{Transferring to different target sensor positions and activities:}\quad 

\noindent
We considered two on-body locations for our source datasets: waist-based Mobiact and wrist-based HHAR.
For Mobiact, we observe effective performance for all target sensor locations including wrist, waist, and ankle. 
However, we  observe that when using Mobiact as source dataset, Waist $\rightarrow$ Wrist and Waist $\rightarrow$ Waist transfer is more successful than Waist $\rightarrow$ Ankle transfer. 
When the sensor location is the same as for Mobiact, i.e., the waist, and for Motionsense and RealWorld, we observe  significant improvements over previous methods. 
In addition, using HHAR as the source also results in performance increases while transferring to different sensor locations.
HHAR transfers better to MHEALTH in contrast to PAMAP2, even though both target domains share the same sensor position (ankle).
The performance gain for waist-based datasets, RealWorld, Motionsense and Mobiact is also better than wrist-based Myogym. 

Therefore, we note that our transfer learning method, Cross-Domain HAR is able to utilize source datasets with varied size, diversity, and sensor positions to transfer across different target sensor locations. 
We obtain similar performance  on MHEALTH, Motionsense and Myogym while using either waist-based Mobiact or wrist-based HHAR as the source, indicating the capability of Cross-Domain HAR to bridge domain gap introduced by differing sensor positions.

Studying the transfer to unseen activities, we note that using Mobiact as source in  Cross-Domain HAR achieves significant boosts of recognition performance for HHAR,  MHEALTH, and Myogym even though the target activities are diverse and present varying degrees of domain gap from source activities.  
Motionsense, which was recorded at the same location and also includes similar activities as Mobiact, gains substantially w.r.t.\ recognition performance.
HHAR as source also gives good performance for Mobiact and RealWorld, which  covers a large number of unseen activities.
In the case of  HHAR transfer to PAMAP2, the performance of Cross-Domain HAR is not good compared to CPC.
We note for the case of Myogym, which covers fine-grained gym activities, Cross-Domain HAR achieves average performance when transferring from source dataset (HHAR) with locomotion style activities.
Overall, if either activities are unseen but the on-body location is the same, or if activities are the same but the on-body location is different, the performance improvements can be larger than when both activities are unseen as well as location is different.

\subsection{Investigating Components of Cross-Domain HAR} 
Previously, we established the superior performance of our method for limited labeled data conditions, relative to diverse baselines across self-supervised and transfer learning. 
We now perform an investigation into the components of our method and study how they contribute to the superior performance. We investigate :
\emph{(i)} the importance of adding the self-supervised loss on unlabeled target data during Student model training; 
\emph{(ii)} the choice of base encoder used in the pipeline ;
\emph{(iii)} the impact of using augmented source data and 
\emph{(iv)} the impact of including consistency regularization in Student Model training. For these experiments, we utilize Mobiact as source dataset. 
We present the results of last two ablation studies i.e. the impact of using source data augmentation in Appendix section \ref{sec:source_data_augmentation} and the impact of including consistency regularization in Appendix section \ref{sec:analysis_consist} respectively.

\subsubsection{Effect of using the self-supervised loss during Student model training}\label{sec:analysis_selfsup}
\begin{figure*}[!t]
    \centering
    \includegraphics[ width=1.0\linewidth]{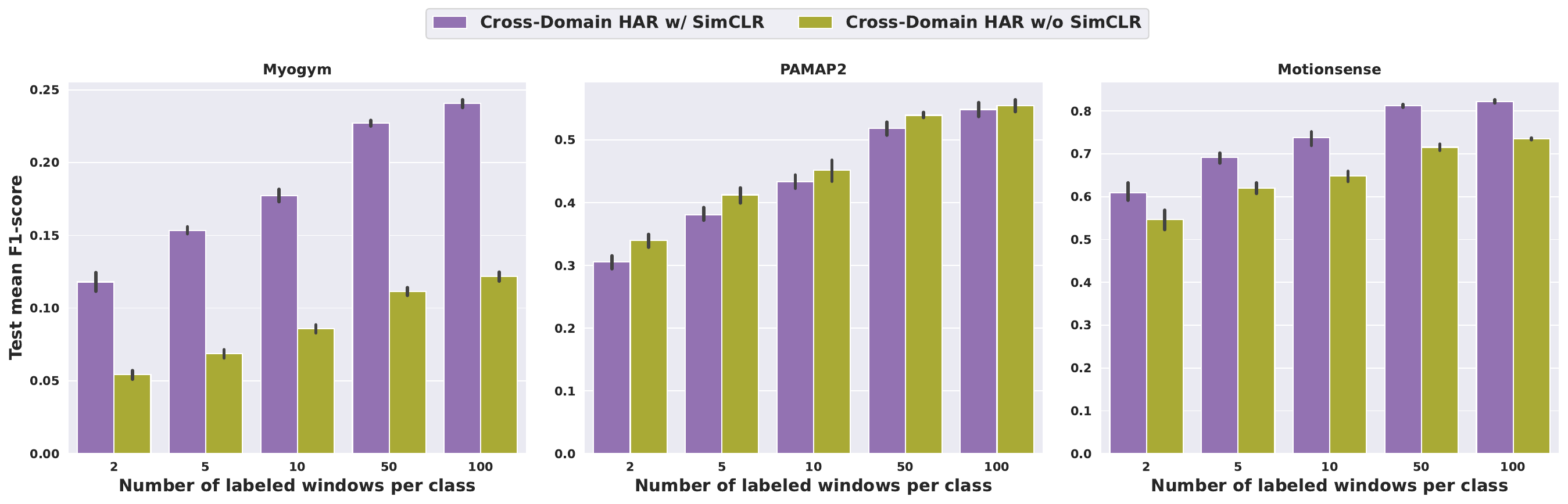}
    \caption{
    Effect of using self-supervised loss during Student model training (using Mobiact source dataset):  
    we study whether the addition of SimCLR-based self-supervision has a positive impact on recognition, when utilized during self-training. 
    We observe that Cross-Domain HAR w/o SimCLR performs consistently lower on most target datasets. 
    For brevity, we only show main results on three target datasets here. 
    Full results are given in Fig.\ \ref{fig:smobi_noss_results} in the Appendix.
    }
    \label{fig:smaller_smobi_noss_results}
\end{figure*}

In this section, we investigate  the need for the self-supervised loss during Student model training.
For self-supervision, we utilized SimCLR \cite{tang2020exploring} as it was shown to be generally highly effective across diverse target scenarios \cite{haresamudram2022assessing}.
Therefore, we visualize the performance of Cross-Domain HAR with and without the self-supervised SimCLR loss in Student model training. 
We denote this as `Cross-Domain HAR w/o SimCLR' in Fig.\ \ref{fig:smaller_smobi_noss_results} and Fig.\ \ref{fig:smobi_noss_results} (in the Appendix) and compare its performance relative to our original setup.

In Fig. \ref{fig:smobi_noss_results}, we observe that the performance without SimCLR loss is  lower for five of six  target datasets especially for $>10$ labeled windows/class of activity data.
Regardless of on-body location or activities, for most datasets including  MHEALTH, HHAR and RealWorld, the performance reduction without the self-supervised loss is around 2-3\%.
For Myogym and  Motionsense in Fig. \ref{fig:smaller_smobi_noss_results},  the impact is significantly stronger, with drops of approx.\ 10\% and 5\%, respectively. 
We postulated in Sec.\ \ref{student_training} that adding self-supervision loss on target data assists in learning additional information from  target domains, that are not captured by the source classes alone, especially when large domain gaps are present from the source dataset.  
For Myogym (where the domain gap is large relative to Mobiact), we note that without using self-supervision the performance is not much better than  Naive transfer (see Fig. \ref{fig:smobi_results}).
This shows that adding the self-supervised loss (i.e., SimCLR) to the self-training setup assists our method to push performance above naively transferring to target domains.

\subsubsection{Impact of the choice of encoder}
\begin{figure*}[!t]
    \centering
    \includegraphics[ width=1.0\linewidth]{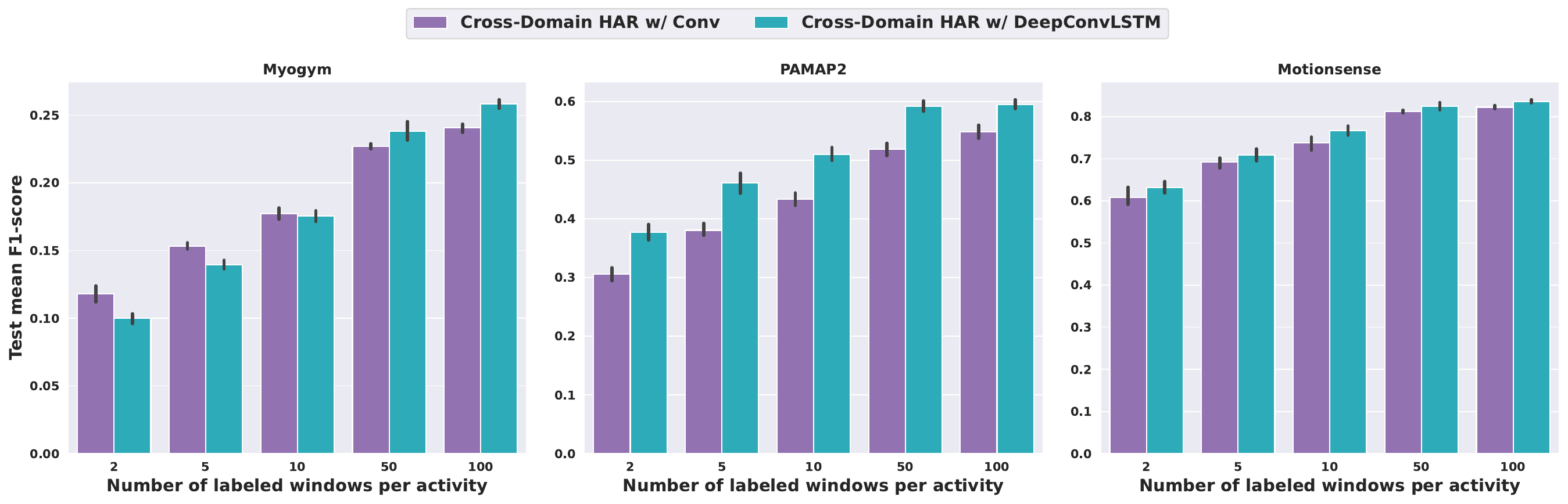}
    \caption{
    Impact of using different encoders in Cross-Domain HAR (using Mobiact source dataset) : we replace the convolutional encoder with DeepConvLSTM \cite{ordonez2016deep}. 
    For four out six target datasets including PAMAP2 and Motionsense, DeepConvLSTM performs better or comparable to convolutional encoder. 
    The choice of encoder has stronger impact when the number of labeled windows/class is lower, as seen for HHAR, PAMAP2 and  MHEALTH. 
    For brevity, we only show main results on three target datasets here. 
    Full results are given in Fig.\ \ref{fig:convrec_results} in the Appendix.
    }
    \label{fig:smaller_convrec_results}
\end{figure*}

In Cross-domain HAR, all stages utilize a common convolutional encoder. 
First, it is used to learn features from a source dataset during Teacher model training, and subsequently  utilized  to adapt and fine-tune the features to the target domain. 
Here, we aim  to analyze whether the encoder architecture should be driven by performance of the encoder on the source domain (i.e., maximizing the Teacher model performance), or on the target domain datasets (i.e., aiming to maximize the final performance on the target). 
\cite{haresamudram2022assessing} studied various model architectures for different datasets -- we refer to their results for comparing performance of different encoders in our pipeline.
We utilize DeepConvLSTM \cite{ordonez2016deep} as the base encoder in Cross-Domain HAR, and train all stages. 
This is referred to as Cross-Domain HAR w/ DeepConvLSTM  in Fig.\ \ref{fig:smaller_convrec_results} and Fig. \ref{fig:convrec_results}. 

For reference, DeepConvLSTM is the more effective classifier for the source dataset (Mobiact) compared to the Conv. classifier, as also shown in \cite{haresamudram2022assessing}. 
We study whether superior performance during Teacher training results in better downstream performance under limited annotation conditions.
For end-to-end training on target datasets (i.e., the full train split), DeepConvLSTM is the better option for Myogym  than the Conv. Classifier (as per \cite{haresamudram2022assessing}).
For rest of the datasets, the Conv. Classifier is found to be a  better option \cite{haresamudram2022assessing}.

From Fig.\ \ref{fig:smaller_convrec_results}, we find that  DeepConvLSTM shows better  performance for Motionsense than the Conv. \ Classifier. 
Similarly, in the case of PAMAP2 as well as MHEALTH and HHAR (Fig.\ \ref{fig:convrec_results} in the
Appendix), we get significantly better recognition performance using DeepConvLSTM as the encoder.
However, in the case of Myogym, the Conv.\ Classifier performs better when the number of labeled windows/class is lower even though DeepConvLSTM is more suitable for both Myogym as well as the source (Mobiact) for end-to-end training (as seen in \cite{haresamudram2022assessing}).
For RealWorld datset, Conv.\ Classifier performs better than DeepConvLSTM across all configurations of number of labeled windows/class. 
For four of six target datasets, DeepConvLSTM is clearly a better encoder choice. 
From these results, it can be hypothesized  that the choice of encoder architecture should be driven more by its performance  on the source domain (i.e., maximizing the Teacher model performance).  
In addition, we observe that the impact of encoder is higher when the number of labeled windows/class is lower (e.g., $\leq5$), as seen for HHAR and MHEALTH in Fig.\ \ref{fig:convrec_results} and  PAMAP2 in Fig.\ \ref{fig:smaller_convrec_results}.

As DeepConvLSTM gives much better performance relative to simple convolutional encoder for most target datasets, it is likely that more powerful encoders, e.g., Transformers \cite{vaswani2017attention} may lead to better performance, although they need to be monitored for overfitting in the few shot setup.

\section{Discussion}
\label{sec:analysis}

Cross-Domain HAR assists in alleviating the impact of lack of sufficient, annotated data, and is highly effective for diverse application scenarios as shown in results in Fig. \ref{fig:smobi_results}.
For example, with just 100 labeled windows / class from the PAMAP2 datset ($\sim$ 5-6\% of data), Cross-Domain HAR achieves $~54$\% F1 score, obtaining performance close to the best performance as obtained by end-to-end training using the full train set ($\sim$60\%) \cite{haresamudram2022assessing}.
Similarly, for RealWorld and Motionsense, Cross-Domain HAR achieves high performance, obtaining approx.\ 78\% and 81\% respectively, enabling the rapid development of activity recognition systems for new target applications.

In particular, the performance is achieved by only utilizing approx.\  two minutes of labeling effort, which is both reasonable and practical.
By utilizing existing labeled HAR datasets and unlabeled target domain data, our method tackles scenarios where well-curated datasets are not available -- a well-known issue in mobile and ubiquitous computing.
Further, it makes maximal use of all available data by not only performing self-training, but also complementing it with self-supervision, such that target domain specific characteristics can be better captured in the representations.

We now summarize our findings for some components of our method and discuss potential follow-up research:

\begin{description}
    \item [Choice of encoder:] Our current setup uses a simple  convolutional encoder for all three components of the framework. 
    As mentioned previously, depending on the relative size of datasets, along with their complexity and underlying activities, different encoders such as DeepConvLSTM and Transformers \cite{vaswani2017attention} can be easily incorporated into our current pipeline for improved performance.

    \item [Using a Combination of Self-training and Self-Supervised Learning:]
    We found in Sec.\ \ref{sec:analysis_selfsup} that the addition of SimCLR-based self-supervision to the self-training setup achieves better recognition.
    This has been previously observed in \cite{xu2021self}, which demonstrates that self-training and pre-training (i.e., self-supervision) are complementary, and can be used for transferring to novel applications with significant domain gaps.
    Combining these semi-supervised approaches is essential as this can assist in making maximal use of both existing labeled and unlabeled datasets to build recognition models for novel application areas with low label availability. 
    In addition, we currently use SimCLR, but other self-supervised methods such as Multi-task self-supervision \cite{saeed2019multi}, Contrastive Predictive Coding \cite{haresamudram2021contrastive}, and SimSiam\cite{chen2021exploring} etc., which may be particularly suited for specific target applications can potentially drive better performance\cite{haresamudram2022assessing}.

    \item[Data Augmentation:]
    Publicly available, labeled datasets for HAR are typically small in size and lack in diversity, which can lead to over-fitting, less generalized features, and simpler model architectures \cite{haresamudram2022assessing}. 
    Transferring such features can be challenging as they can be specific to the source domain. 
    We find that applying augmentation transformations increases the source datasets' size and underlying data diversity, helping us improve performance for supervised learning as well as self-training methods. 
    We demonstrate this in Sec.\ \ref{sec:source_data_augmentation}, where we observe superior performance by using data augmentations, indicating that the learned features can better transfer to diverse target domains.
    Given the importance of augmentations, developing automatic augmentation strategies as in \cite{zoph2020learning, cubuk2019autoaugment} specifically for self-training could lead to automatically deriving the answer to which augmentation stratagies are better.
\end{description}

\paragraph{Limitations and Future Work}
In our experiments, we chose the source labeled dataset without explicitly considering its suitability for target domains. 
Although we analyze how well Mobiact/HHAR can perform as the source, our analysis mainly focuses on their relative sizes and underlying diversity.
Given access to a number of labeled source datasets, a potential research direction involves developing techniques that can indicate the suitability of source datasets for specific target applications, especially without any / with minimal additional training.

Additionally, the tasks presented to the Teacher and Student models have different levels of difficulty, in form of access to weakly and strongly augmented data from the target.
Therefore, examining which augmentations are most impactful for wearable sensor data, and which combination of task difficulties (weak-weak, weak-strong, strong-weak, strong-strong) can lead to further improvements, as in   \cite{islam2021dynamic}.
Finally, our setup assumes access to unlabeled data from target domains, which may not be easy to collect.
Therefore, analyzing the minimum quantities of unlabeled target data required for our setup could have  practical implications. 
It can assist in reducing the need of resources from application scenarios while  achieving effective performance for recognition.

\section{Conclusion}
Building robust and reliable activity recognition systems generally requires significant amounts of labeled data, especially when relying on supervised deep learning.
The HAR community is plagued with the issue of lack of well-labeled, large-scale datasets, as reliable ground truth annotation is difficult to obtain \cite{saeed2019multi, kwon2020imutube}. 
For example, making sense of signal data is not as intuitive  as for modalities like images or text, and can require domain-expertise, making it expensive and time-consuming \cite{tang2021selfhar}.
In addition, user-related privacy issues also pose additional constraints for large-scale collection \cite{saeed2019multi}.

These challenges have motivated the community to leverage and deploy alternate methods such that the reliance on costly annotations can be reduced.
Many recent works have proposed self-supervised methods \cite{haresamudram2020masked, saeed2019multi,tang2020exploring} which leverage the unlabeled sensor data from smartphones and smartwatches to learn effective representations.
Our work is in line with such data-efficient methods, so as to activity recognition systems without the need of large-scale annotations from the target domain.

Our approach differs from self-supervision as it instead leverages existing labeled datasets for pre-training, as well as unlabeled target data to build reliable activity recognition systems.
The key idea is to leverage the models trained using source domain data to pseudo-label target data and use it for training.
To better adapt the representations to a different domain, we employ a number of modifications incl.\ source data augmentation, consistency regularization \cite{hady2013semi}, and the addition of self-supervision to Student model training.
Finally, the Student model weights are fine-tuned to the target domain using very limited labeled windows of target data.

For few shot settings, Cross-Domain HAR outperforms self-supervised as well as end-to-end methods across target conditions, \edit{and is capable of leveraging source domains with varying end activities, diversity and dataset sizes}.
These results are promising as they show that an existing labeled datasets can be instrumental in building recognition systems for a large number of downstream applications, while using a small fraction of annotation effort needed compared to supervised learning.
Overall, our work adds to the push for reducing reliance on annotations, and towards  developing robust models while expending minimal annotation effort.





\bibliographystyle{ACM-Reference-Format}
\bibliography{refs}


\begin{thebibliography}{57}


\ifx \showCODEN    \undefined \def \showCODEN     #1{\unskip}     \fi
\ifx \showDOI      \undefined \def \showDOI       #1{#1}\fi
\ifx \showISBNx    \undefined \def \showISBNx     #1{\unskip}     \fi
\ifx \showISBNxiii \undefined \def \showISBNxiii  #1{\unskip}     \fi
\ifx \showISSN     \undefined \def \showISSN      #1{\unskip}     \fi
\ifx \showLCCN     \undefined \def \showLCCN      #1{\unskip}     \fi
\ifx \shownote     \undefined \def \shownote      #1{#1}          \fi
\ifx \showarticletitle \undefined \def \showarticletitle #1{#1}   \fi
\ifx \showURL      \undefined \def \showURL       {\relax}        \fi
\providecommand\bibfield[2]{#2}
\providecommand\bibinfo[2]{#2}
\providecommand\natexlab[1]{#1}
\providecommand\showeprint[2][]{arXiv:#2}

\bibitem[Altun et~al\mbox{.}(2010)]%
        {altun2010comparative}
\bibfield{author}{\bibinfo{person}{Kerem Altun}, \bibinfo{person}{Billur Barshan}, {and} \bibinfo{person}{Orkun Tun{\c{c}}el}.} \bibinfo{year}{2010}\natexlab{}.
\newblock \showarticletitle{Comparative study on classifying human activities with miniature inertial and magnetic sensors}.
\newblock \bibinfo{journal}{\emph{Pattern Recognition}} \bibinfo{volume}{43}, \bibinfo{number}{10} (\bibinfo{year}{2010}), \bibinfo{pages}{3605--3620}.
\newblock


\bibitem[An et~al\mbox{.}(2020)]%
        {an2020transfer}
\bibfield{author}{\bibinfo{person}{Sizhe An}, \bibinfo{person}{Ganapati Bhat}, \bibinfo{person}{Suat Gumussoy}, {and} \bibinfo{person}{Umit Ogras}.} \bibinfo{year}{2020}\natexlab{}.
\newblock \showarticletitle{Transfer learning for human activity recognition using representational analysis of neural networks}.
\newblock \bibinfo{journal}{\emph{arXiv preprint arXiv:2012.04479}} (\bibinfo{year}{2020}).
\newblock


\bibitem[Bachlin et~al\mbox{.}(2009)]%
        {bachlin2009wearable}
\bibfield{author}{\bibinfo{person}{Marc Bachlin}, \bibinfo{person}{Meir Plotnik}, \bibinfo{person}{Daniel Roggen}, \bibinfo{person}{Inbal Maidan}, \bibinfo{person}{Jeffrey~M Hausdorff}, \bibinfo{person}{Nir Giladi}, {and} \bibinfo{person}{Gerhard Troster}.} \bibinfo{year}{2009}\natexlab{}.
\newblock \showarticletitle{Wearable assistant for Parkinson’s disease patients with the freezing of gait symptom}.
\newblock \bibinfo{journal}{\emph{IEEE Transactions on Information Technology in Biomedicine}} \bibinfo{volume}{14}, \bibinfo{number}{2} (\bibinfo{year}{2009}), \bibinfo{pages}{436--446}.
\newblock


\bibitem[Banos et~al\mbox{.}(2014)]%
        {banos2014mhealthdroid}
\bibfield{author}{\bibinfo{person}{Oresti Banos}, \bibinfo{person}{Rafael Garcia}, \bibinfo{person}{Juan~A Holgado-Terriza}, \bibinfo{person}{Miguel Damas}, \bibinfo{person}{Hector Pomares}, \bibinfo{person}{Ignacio Rojas}, \bibinfo{person}{Alejandro Saez}, {and} \bibinfo{person}{Claudia Villalonga}.} \bibinfo{year}{2014}\natexlab{}.
\newblock \showarticletitle{mHealthDroid: a novel framework for agile development of mobile health applications}. In \bibinfo{booktitle}{\emph{International workshop on ambient assisted living}}. Springer, \bibinfo{pages}{91--98}.
\newblock


\bibitem[Bin~Morshed et~al\mbox{.}(2020)]%
        {bin2020measuring}
\bibfield{author}{\bibinfo{person}{Mehrab Bin~Morshed}, \bibinfo{person}{Koustuv Saha}, \bibinfo{person}{Munmun De~Choudhury}, \bibinfo{person}{Gregory~D Abowd}, {and} \bibinfo{person}{Thomas Pl{\"o}tz}.} \bibinfo{year}{2020}\natexlab{}.
\newblock \showarticletitle{Measuring self-esteem with passive sensing}. In \bibinfo{booktitle}{\emph{Proceedings of the 14th EAI International Conference on Pervasive Computing Technologies for Healthcare}}. \bibinfo{pages}{363--366}.
\newblock


\bibitem[Chatzaki et~al\mbox{.}(2016)]%
        {chatzaki2016human}
\bibfield{author}{\bibinfo{person}{Charikleia Chatzaki}, \bibinfo{person}{Matthew Pediaditis}, \bibinfo{person}{George Vavoulas}, {and} \bibinfo{person}{Manolis Tsiknakis}.} \bibinfo{year}{2016}\natexlab{}.
\newblock \showarticletitle{Human daily activity and fall recognition using a smartphone’s acceleration sensor}. In \bibinfo{booktitle}{\emph{International Conference on Information and Communication Technologies for Ageing Well and e-Health}}. Springer, \bibinfo{pages}{100--118}.
\newblock


\bibitem[Chen et~al\mbox{.}(2020)]%
        {chen2020simple}
\bibfield{author}{\bibinfo{person}{Ting Chen}, \bibinfo{person}{Simon Kornblith}, \bibinfo{person}{Mohammad Norouzi}, {and} \bibinfo{person}{Geoffrey Hinton}.} \bibinfo{year}{2020}\natexlab{}.
\newblock \showarticletitle{A simple framework for contrastive learning of visual representations}. In \bibinfo{booktitle}{\emph{International conference on machine learning}}. PMLR, \bibinfo{pages}{1597--1607}.
\newblock


\bibitem[Chen and He(2021)]%
        {chen2021exploring}
\bibfield{author}{\bibinfo{person}{Xinlei Chen} {and} \bibinfo{person}{Kaiming He}.} \bibinfo{year}{2021}\natexlab{}.
\newblock \showarticletitle{Exploring simple siamese representation learning}. In \bibinfo{booktitle}{\emph{Proceedings of the IEEE/CVF conference on computer vision and pattern recognition}}. \bibinfo{pages}{15750--15758}.
\newblock


\bibitem[Cook et~al\mbox{.}(2013)]%
        {cook2013transfer}
\bibfield{author}{\bibinfo{person}{Diane Cook}, \bibinfo{person}{Kyle~D Feuz}, {and} \bibinfo{person}{Narayanan~C Krishnan}.} \bibinfo{year}{2013}\natexlab{}.
\newblock \showarticletitle{Transfer learning for activity recognition: A survey}.
\newblock \bibinfo{journal}{\emph{Knowledge and information systems}}  \bibinfo{volume}{36} (\bibinfo{year}{2013}), \bibinfo{pages}{537--556}.
\newblock


\bibitem[Cubuk et~al\mbox{.}(2019)]%
        {cubuk2019autoaugment}
\bibfield{author}{\bibinfo{person}{Ekin~D Cubuk}, \bibinfo{person}{Barret Zoph}, \bibinfo{person}{Dandelion Mane}, \bibinfo{person}{Vijay Vasudevan}, {and} \bibinfo{person}{Quoc~V Le}.} \bibinfo{year}{2019}\natexlab{}.
\newblock \showarticletitle{Autoaugment: Learning augmentation strategies from data}. In \bibinfo{booktitle}{\emph{Proceedings of the IEEE/CVF conference on computer vision and pattern recognition}}. \bibinfo{pages}{113--123}.
\newblock


\bibitem[Deng et~al\mbox{.}(2009)]%
        {deng2009imagenet}
\bibfield{author}{\bibinfo{person}{Jia Deng}, \bibinfo{person}{Wei Dong}, \bibinfo{person}{Richard Socher}, \bibinfo{person}{Li-Jia Li}, \bibinfo{person}{Kai Li}, {and} \bibinfo{person}{Li Fei-Fei}.} \bibinfo{year}{2009}\natexlab{}.
\newblock \showarticletitle{Imagenet: A large-scale hierarchical image database}. In \bibinfo{booktitle}{\emph{2009 IEEE conference on computer vision and pattern recognition}}. Ieee, \bibinfo{pages}{248--255}.
\newblock


\bibitem[Ding et~al\mbox{.}(2018)]%
        {ding2018empirical}
\bibfield{author}{\bibinfo{person}{Renjie Ding}, \bibinfo{person}{Xue Li}, \bibinfo{person}{Lanshun Nie}, \bibinfo{person}{Jiazhen Li}, \bibinfo{person}{Xiandong Si}, \bibinfo{person}{Dianhui Chu}, \bibinfo{person}{Guozhong Liu}, {and} \bibinfo{person}{Dechen Zhan}.} \bibinfo{year}{2018}\natexlab{}.
\newblock \showarticletitle{Empirical study and improvement on deep transfer learning for human activity recognition}.
\newblock \bibinfo{journal}{\emph{Sensors}} \bibinfo{volume}{19}, \bibinfo{number}{1} (\bibinfo{year}{2018}), \bibinfo{pages}{57}.
\newblock


\bibitem[Ericsson et~al\mbox{.}(2022)]%
        {ericsson2022self}
\bibfield{author}{\bibinfo{person}{Linus Ericsson}, \bibinfo{person}{Henry Gouk}, \bibinfo{person}{Chen~Change Loy}, {and} \bibinfo{person}{Timothy~M Hospedales}.} \bibinfo{year}{2022}\natexlab{}.
\newblock \showarticletitle{Self-supervised representation learning: Introduction, advances, and challenges}.
\newblock \bibinfo{journal}{\emph{IEEE Signal Processing Magazine}} \bibinfo{volume}{39}, \bibinfo{number}{3} (\bibinfo{year}{2022}), \bibinfo{pages}{42--62}.
\newblock


\bibitem[Finn et~al\mbox{.}(2017)]%
        {finn2017model}
\bibfield{author}{\bibinfo{person}{Chelsea Finn}, \bibinfo{person}{Pieter Abbeel}, {and} \bibinfo{person}{Sergey Levine}.} \bibinfo{year}{2017}\natexlab{}.
\newblock \showarticletitle{Model-agnostic meta-learning for fast adaptation of deep networks}. In \bibinfo{booktitle}{\emph{International conference on machine learning}}. PMLR, \bibinfo{pages}{1126--1135}.
\newblock


\bibitem[Hady and Schwenker(2013)]%
        {hady2013semi}
\bibfield{author}{\bibinfo{person}{Mohamed Farouk~Abdel Hady} {and} \bibinfo{person}{Friedhelm Schwenker}.} \bibinfo{year}{2013}\natexlab{}.
\newblock \showarticletitle{Semi-supervised learning}.
\newblock \bibinfo{journal}{\emph{Handbook on Neural Information Processing}} (\bibinfo{year}{2013}), \bibinfo{pages}{215--239}.
\newblock


\bibitem[Hammerla et~al\mbox{.}(2016)]%
        {hammerla2016deep}
\bibfield{author}{\bibinfo{person}{Nils~Y Hammerla}, \bibinfo{person}{Shane Halloran}, {and} \bibinfo{person}{Thomas Pl{\"o}tz}.} \bibinfo{year}{2016}\natexlab{}.
\newblock \showarticletitle{Deep, convolutional, and recurrent models for human activity recognition using wearables}.
\newblock \bibinfo{journal}{\emph{arXiv preprint arXiv:1604.08880}} (\bibinfo{year}{2016}).
\newblock


\bibitem[Haresamudram et~al\mbox{.}(2020)]%
        {haresamudram2020masked}
\bibfield{author}{\bibinfo{person}{Harish Haresamudram}, \bibinfo{person}{Apoorva Beedu}, \bibinfo{person}{Varun Agrawal}, \bibinfo{person}{Patrick~L Grady}, \bibinfo{person}{Irfan Essa}, \bibinfo{person}{Judy Hoffman}, {and} \bibinfo{person}{Thomas Pl{\"o}tz}.} \bibinfo{year}{2020}\natexlab{}.
\newblock \showarticletitle{Masked reconstruction based self-supervision for human activity recognition}. In \bibinfo{booktitle}{\emph{Proceedings of the 2020 ACM International Symposium on Wearable Computers}}. \bibinfo{pages}{45--49}.
\newblock


\bibitem[Haresamudram et~al\mbox{.}(2021)]%
        {haresamudram2021contrastive}
\bibfield{author}{\bibinfo{person}{Harish Haresamudram}, \bibinfo{person}{Irfan Essa}, {and} \bibinfo{person}{Thomas Pl{\"o}tz}.} \bibinfo{year}{2021}\natexlab{}.
\newblock \showarticletitle{Contrastive predictive coding for human activity recognition}.
\newblock \bibinfo{journal}{\emph{Proceedings of the ACM on Interactive, Mobile, Wearable and Ubiquitous Technologies}} \bibinfo{volume}{5}, \bibinfo{number}{2} (\bibinfo{year}{2021}), \bibinfo{pages}{1--26}.
\newblock


\bibitem[Haresamudram et~al\mbox{.}(2022)]%
        {haresamudram2022assessing}
\bibfield{author}{\bibinfo{person}{Harish Haresamudram}, \bibinfo{person}{Irfan Essa}, {and} \bibinfo{person}{Thomas Pl{\"o}tz}.} \bibinfo{year}{2022}\natexlab{}.
\newblock \showarticletitle{Assessing the state of self-supervised human activity recognition using wearables}.
\newblock \bibinfo{journal}{\emph{Proceedings of the ACM on Interactive, Mobile, Wearable and Ubiquitous Technologies}} \bibinfo{volume}{6}, \bibinfo{number}{3} (\bibinfo{year}{2022}), \bibinfo{pages}{1--47}.
\newblock


\bibitem[Hinton et~al\mbox{.}(2015)]%
        {hinton2015distilling}
\bibfield{author}{\bibinfo{person}{Geoffrey Hinton}, \bibinfo{person}{Oriol Vinyals}, {and} \bibinfo{person}{Jeff Dean}.} \bibinfo{year}{2015}\natexlab{}.
\newblock \showarticletitle{Distilling the knowledge in a neural network}.
\newblock \bibinfo{journal}{\emph{arXiv preprint arXiv:1503.02531}} (\bibinfo{year}{2015}).
\newblock


\bibitem[Hoelzemann and Van~Laerhoven(2020)]%
        {hoelzemann2020digging}
\bibfield{author}{\bibinfo{person}{Alexander Hoelzemann} {and} \bibinfo{person}{Kristof Van~Laerhoven}.} \bibinfo{year}{2020}\natexlab{}.
\newblock \showarticletitle{Digging deeper: towards a better understanding of transfer learning for human activity recognition}. In \bibinfo{booktitle}{\emph{Proceedings of the 2020 International Symposium on Wearable Computers}}. \bibinfo{pages}{50--54}.
\newblock


\bibitem[Ioffe and Szegedy(2015)]%
        {ioffe2015batch}
\bibfield{author}{\bibinfo{person}{Sergey Ioffe} {and} \bibinfo{person}{Christian Szegedy}.} \bibinfo{year}{2015}\natexlab{}.
\newblock \showarticletitle{Batch normalization: Accelerating deep network training by reducing internal covariate shift}. In \bibinfo{booktitle}{\emph{International conference on machine learning}}. pmlr, \bibinfo{pages}{448--456}.
\newblock


\bibitem[Islam et~al\mbox{.}(2021)]%
        {islam2021dynamic}
\bibfield{author}{\bibinfo{person}{Ashraful Islam}, \bibinfo{person}{Chun-Fu~Richard Chen}, \bibinfo{person}{Rameswar Panda}, \bibinfo{person}{Leonid Karlinsky}, \bibinfo{person}{Rogerio Feris}, {and} \bibinfo{person}{Richard~J Radke}.} \bibinfo{year}{2021}\natexlab{}.
\newblock \showarticletitle{Dynamic distillation network for cross-domain few-shot recognition with unlabeled data}.
\newblock \bibinfo{journal}{\emph{Advances in Neural Information Processing Systems}}  \bibinfo{volume}{34} (\bibinfo{year}{2021}), \bibinfo{pages}{3584--3595}.
\newblock


\bibitem[Jain et~al\mbox{.}(2022)]%
        {jain2022collossl}
\bibfield{author}{\bibinfo{person}{Yash Jain}, \bibinfo{person}{Chi~Ian Tang}, \bibinfo{person}{Chulhong Min}, \bibinfo{person}{Fahim Kawsar}, {and} \bibinfo{person}{Akhil Mathur}.} \bibinfo{year}{2022}\natexlab{}.
\newblock \showarticletitle{ColloSSL: Collaborative self-supervised learning for human activity recognition}.
\newblock \bibinfo{journal}{\emph{Proceedings of the ACM on Interactive, Mobile, Wearable and Ubiquitous Technologies}} \bibinfo{volume}{6}, \bibinfo{number}{1} (\bibinfo{year}{2022}), \bibinfo{pages}{1--28}.
\newblock


\bibitem[Koskim{\"a}ki et~al\mbox{.}(2017)]%
        {koskimaki2017myogym}
\bibfield{author}{\bibinfo{person}{Heli Koskim{\"a}ki}, \bibinfo{person}{Pekka Siirtola}, {and} \bibinfo{person}{Juha R{\"o}ning}.} \bibinfo{year}{2017}\natexlab{}.
\newblock \showarticletitle{Myogym: introducing an open gym data set for activity recognition collected using myo armband}. In \bibinfo{booktitle}{\emph{Proceedings of the 2017 ACM International Joint Conference on Pervasive and Ubiquitous Computing and Proceedings of the 2017 ACM International Symposium on Wearable Computers}}. \bibinfo{pages}{537--546}.
\newblock


\bibitem[Kwon et~al\mbox{.}(2020)]%
        {kwon2020imutube}
\bibfield{author}{\bibinfo{person}{Hyeokhyen Kwon}, \bibinfo{person}{Catherine Tong}, \bibinfo{person}{Harish Haresamudram}, \bibinfo{person}{Yan Gao}, \bibinfo{person}{Gregory~D Abowd}, \bibinfo{person}{Nicholas~D Lane}, {and} \bibinfo{person}{Thomas Ploetz}.} \bibinfo{year}{2020}\natexlab{}.
\newblock \showarticletitle{IMUTube: Automatic extraction of virtual on-body accelerometry from video for human activity recognition}.
\newblock \bibinfo{journal}{\emph{Proceedings of the ACM on Interactive, Mobile, Wearable and Ubiquitous Technologies}} \bibinfo{volume}{4}, \bibinfo{number}{3} (\bibinfo{year}{2020}), \bibinfo{pages}{1--29}.
\newblock


\bibitem[Lara and Labrador(2012)]%
        {lara2012survey}
\bibfield{author}{\bibinfo{person}{Oscar~D Lara} {and} \bibinfo{person}{Miguel~A Labrador}.} \bibinfo{year}{2012}\natexlab{}.
\newblock \showarticletitle{A survey on human activity recognition using wearable sensors}.
\newblock \bibinfo{journal}{\emph{IEEE communications surveys \& tutorials}} \bibinfo{volume}{15}, \bibinfo{number}{3} (\bibinfo{year}{2012}), \bibinfo{pages}{1192--1209}.
\newblock


\bibitem[Malekzadeh et~al\mbox{.}(2018)]%
        {malekzadeh2018protecting}
\bibfield{author}{\bibinfo{person}{Mohammad Malekzadeh}, \bibinfo{person}{Richard~G Clegg}, \bibinfo{person}{Andrea Cavallaro}, {and} \bibinfo{person}{Hamed Haddadi}.} \bibinfo{year}{2018}\natexlab{}.
\newblock \showarticletitle{Protecting sensory data against sensitive inferences}. In \bibinfo{booktitle}{\emph{Proceedings of the 1st Workshop on Privacy by Design in Distributed Systems}}. \bibinfo{pages}{1--6}.
\newblock


\bibitem[Morales and Roggen(2016)]%
        {morales2016deep}
\bibfield{author}{\bibinfo{person}{Francisco Javier~Ord{\'o}{\~n}ez Morales} {and} \bibinfo{person}{Daniel Roggen}.} \bibinfo{year}{2016}\natexlab{}.
\newblock \showarticletitle{Deep convolutional feature transfer across mobile activity recognition domains, sensor modalities and locations}. In \bibinfo{booktitle}{\emph{Proceedings of the 2016 ACM International Symposium on Wearable Computers}}. \bibinfo{pages}{92--99}.
\newblock


\bibitem[Morshed et~al\mbox{.}(2019)]%
        {morshed2019prediction}
\bibfield{author}{\bibinfo{person}{Mehrab~Bin Morshed}, \bibinfo{person}{Koustuv Saha}, \bibinfo{person}{Richard Li}, \bibinfo{person}{Sidney~K D'Mello}, \bibinfo{person}{Munmun De~Choudhury}, \bibinfo{person}{Gregory~D Abowd}, {and} \bibinfo{person}{Thomas Pl{\"o}tz}.} \bibinfo{year}{2019}\natexlab{}.
\newblock \showarticletitle{Prediction of mood instability with passive sensing}.
\newblock \bibinfo{journal}{\emph{Proceedings of the ACM on Interactive, Mobile, Wearable and Ubiquitous Technologies}} \bibinfo{volume}{3}, \bibinfo{number}{3} (\bibinfo{year}{2019}), \bibinfo{pages}{1--21}.
\newblock


\bibitem[Nair and Hinton(2010)]%
        {nair2010rectified}
\bibfield{author}{\bibinfo{person}{Vinod Nair} {and} \bibinfo{person}{Geoffrey~E Hinton}.} \bibinfo{year}{2010}\natexlab{}.
\newblock \showarticletitle{Rectified linear units improve restricted boltzmann machines}. In \bibinfo{booktitle}{\emph{Proceedings of the 27th international conference on machine learning (ICML-10)}}. \bibinfo{pages}{807--814}.
\newblock


\bibitem[Oord et~al\mbox{.}(2018)]%
        {oord2018representation}
\bibfield{author}{\bibinfo{person}{Aaron van~den Oord}, \bibinfo{person}{Yazhe Li}, {and} \bibinfo{person}{Oriol Vinyals}.} \bibinfo{year}{2018}\natexlab{}.
\newblock \showarticletitle{Representation learning with contrastive predictive coding}.
\newblock \bibinfo{journal}{\emph{arXiv preprint arXiv:1807.03748}} (\bibinfo{year}{2018}).
\newblock


\bibitem[Ord{\'o}{\~n}ez and Roggen(2016)]%
        {ordonez2016deep}
\bibfield{author}{\bibinfo{person}{Francisco~Javier Ord{\'o}{\~n}ez} {and} \bibinfo{person}{Daniel Roggen}.} \bibinfo{year}{2016}\natexlab{}.
\newblock \showarticletitle{Deep convolutional and lstm recurrent neural networks for multimodal wearable activity recognition}.
\newblock \bibinfo{journal}{\emph{Sensors}} \bibinfo{volume}{16}, \bibinfo{number}{1} (\bibinfo{year}{2016}), \bibinfo{pages}{115}.
\newblock


\bibitem[Ouali et~al\mbox{.}(2020)]%
        {ouali2020overview}
\bibfield{author}{\bibinfo{person}{Yassine Ouali}, \bibinfo{person}{C{\'e}line Hudelot}, {and} \bibinfo{person}{Myriam Tami}.} \bibinfo{year}{2020}\natexlab{}.
\newblock \showarticletitle{An overview of deep semi-supervised learning}.
\newblock \bibinfo{journal}{\emph{arXiv preprint arXiv:2006.05278}} (\bibinfo{year}{2020}).
\newblock


\bibitem[Paszke et~al\mbox{.}(2019)]%
        {paszke2019pytorch}
\bibfield{author}{\bibinfo{person}{Adam Paszke}, \bibinfo{person}{Sam Gross}, \bibinfo{person}{Francisco Massa}, \bibinfo{person}{Adam Lerer}, \bibinfo{person}{James Bradbury}, \bibinfo{person}{Gregory Chanan}, \bibinfo{person}{Trevor Killeen}, \bibinfo{person}{Zeming Lin}, \bibinfo{person}{Natalia Gimelshein}, \bibinfo{person}{Luca Antiga}, {et~al\mbox{.}}} \bibinfo{year}{2019}\natexlab{}.
\newblock \showarticletitle{Pytorch: An imperative style, high-performance deep learning library}.
\newblock \bibinfo{journal}{\emph{Advances in neural information processing systems}}  \bibinfo{volume}{32} (\bibinfo{year}{2019}).
\newblock


\bibitem[Phoo and Hariharan(2020)]%
        {phoo2020self}
\bibfield{author}{\bibinfo{person}{Cheng~Perng Phoo} {and} \bibinfo{person}{Bharath Hariharan}.} \bibinfo{year}{2020}\natexlab{}.
\newblock \showarticletitle{Self-training for few-shot transfer across extreme task differences}.
\newblock \bibinfo{journal}{\emph{arXiv preprint arXiv:2010.07734}} (\bibinfo{year}{2020}).
\newblock


\bibitem[Pl{\"O}tz(2021)]%
        {plotz2021applying}
\bibfield{author}{\bibinfo{person}{Thomas Pl{\"O}tz}.} \bibinfo{year}{2021}\natexlab{}.
\newblock \showarticletitle{Applying machine learning for sensor data analysis in interactive systems: Common pitfalls of pragmatic use and ways to avoid them}.
\newblock \bibinfo{journal}{\emph{ACM Computing Surveys (CSUR)}} \bibinfo{volume}{54}, \bibinfo{number}{6} (\bibinfo{year}{2021}), \bibinfo{pages}{1--25}.
\newblock


\bibitem[Powers(2020)]%
        {powers2020evaluation}
\bibfield{author}{\bibinfo{person}{David~MW Powers}.} \bibinfo{year}{2020}\natexlab{}.
\newblock \showarticletitle{Evaluation: from precision, recall and F-measure to ROC, informedness, markedness and correlation}.
\newblock \bibinfo{journal}{\emph{arXiv preprint arXiv:2010.16061}} (\bibinfo{year}{2020}).
\newblock


\bibitem[Reiss and Stricker(2012)]%
        {reiss2012introducing}
\bibfield{author}{\bibinfo{person}{Attila Reiss} {and} \bibinfo{person}{Didier Stricker}.} \bibinfo{year}{2012}\natexlab{}.
\newblock \showarticletitle{Introducing a new benchmarked dataset for activity monitoring}. In \bibinfo{booktitle}{\emph{2012 16th international symposium on wearable computers}}. IEEE, \bibinfo{pages}{108--109}.
\newblock


\bibitem[Saeed et~al\mbox{.}(2019)]%
        {saeed2019multi}
\bibfield{author}{\bibinfo{person}{Aaqib Saeed}, \bibinfo{person}{Tanir Ozcelebi}, {and} \bibinfo{person}{Johan Lukkien}.} \bibinfo{year}{2019}\natexlab{}.
\newblock \showarticletitle{Multi-task self-supervised learning for human activity detection}.
\newblock \bibinfo{journal}{\emph{Proceedings of the ACM on Interactive, Mobile, Wearable and Ubiquitous Technologies}} \bibinfo{volume}{3}, \bibinfo{number}{2} (\bibinfo{year}{2019}), \bibinfo{pages}{1--30}.
\newblock


\bibitem[Snell et~al\mbox{.}(2017)]%
        {snell2017prototypical}
\bibfield{author}{\bibinfo{person}{Jake Snell}, \bibinfo{person}{Kevin Swersky}, {and} \bibinfo{person}{Richard Zemel}.} \bibinfo{year}{2017}\natexlab{}.
\newblock \showarticletitle{Prototypical networks for few-shot learning}.
\newblock \bibinfo{journal}{\emph{Advances in neural information processing systems}}  \bibinfo{volume}{30} (\bibinfo{year}{2017}).
\newblock


\bibitem[Srivastava et~al\mbox{.}(2014)]%
        {srivastava2014dropout}
\bibfield{author}{\bibinfo{person}{Nitish Srivastava}, \bibinfo{person}{Geoffrey Hinton}, \bibinfo{person}{Alex Krizhevsky}, \bibinfo{person}{Ilya Sutskever}, {and} \bibinfo{person}{Ruslan Salakhutdinov}.} \bibinfo{year}{2014}\natexlab{}.
\newblock \showarticletitle{Dropout: a simple way to prevent neural networks from overfitting}.
\newblock \bibinfo{journal}{\emph{The journal of machine learning research}} \bibinfo{volume}{15}, \bibinfo{number}{1} (\bibinfo{year}{2014}), \bibinfo{pages}{1929--1958}.
\newblock


\bibitem[Stisen et~al\mbox{.}(2015)]%
        {stisen2015smart}
\bibfield{author}{\bibinfo{person}{Allan Stisen}, \bibinfo{person}{Henrik Blunck}, \bibinfo{person}{Sourav Bhattacharya}, \bibinfo{person}{Thor~Siiger Prentow}, \bibinfo{person}{Mikkel~Baun Kj{\ae}rgaard}, \bibinfo{person}{Anind Dey}, \bibinfo{person}{Tobias Sonne}, {and} \bibinfo{person}{Mads~M{\o}ller Jensen}.} \bibinfo{year}{2015}\natexlab{}.
\newblock \showarticletitle{Smart devices are different: Assessing and mitigatingmobile sensing heterogeneities for activity recognition}. In \bibinfo{booktitle}{\emph{Proceedings of the 13th ACM conference on embedded networked sensor systems}}. \bibinfo{pages}{127--140}.
\newblock


\bibitem[Sztyler and Stuckenschmidt(2016)]%
        {sztyler2016body}
\bibfield{author}{\bibinfo{person}{Timo Sztyler} {and} \bibinfo{person}{Heiner Stuckenschmidt}.} \bibinfo{year}{2016}\natexlab{}.
\newblock \showarticletitle{On-body localization of wearable devices: An investigation of position-aware activity recognition}. In \bibinfo{booktitle}{\emph{2016 IEEE International Conference on Pervasive Computing and Communications (PerCom)}}. IEEE, \bibinfo{pages}{1--9}.
\newblock


\bibitem[Tang et~al\mbox{.}(2021)]%
        {tang2021selfhar}
\bibfield{author}{\bibinfo{person}{Chi~Ian Tang}, \bibinfo{person}{Ignacio Perez-Pozuelo}, \bibinfo{person}{Dimitris Spathis}, \bibinfo{person}{Soren Brage}, \bibinfo{person}{Nick Wareham}, {and} \bibinfo{person}{Cecilia Mascolo}.} \bibinfo{year}{2021}\natexlab{}.
\newblock \showarticletitle{Selfhar: Improving human activity recognition through self-training with unlabeled data}.
\newblock \bibinfo{journal}{\emph{arXiv preprint arXiv:2102.06073}} (\bibinfo{year}{2021}).
\newblock


\bibitem[Tang et~al\mbox{.}(2020)]%
        {tang2020exploring}
\bibfield{author}{\bibinfo{person}{Chi~Ian Tang}, \bibinfo{person}{Ignacio Perez-Pozuelo}, \bibinfo{person}{Dimitris Spathis}, {and} \bibinfo{person}{Cecilia Mascolo}.} \bibinfo{year}{2020}\natexlab{}.
\newblock \showarticletitle{Exploring contrastive learning in human activity recognition for healthcare}.
\newblock \bibinfo{journal}{\emph{arXiv preprint arXiv:2011.11542}} (\bibinfo{year}{2020}).
\newblock


\bibitem[Um et~al\mbox{.}(2017)]%
        {um2017data}
\bibfield{author}{\bibinfo{person}{Terry~T Um}, \bibinfo{person}{Franz~MJ Pfister}, \bibinfo{person}{Daniel Pichler}, \bibinfo{person}{Satoshi Endo}, \bibinfo{person}{Muriel Lang}, \bibinfo{person}{Sandra Hirche}, \bibinfo{person}{Urban Fietzek}, {and} \bibinfo{person}{Dana Kuli{\'c}}.} \bibinfo{year}{2017}\natexlab{}.
\newblock \showarticletitle{Data augmentation of wearable sensor data for parkinson’s disease monitoring using convolutional neural networks}. In \bibinfo{booktitle}{\emph{Proceedings of the 19th ACM international conference on multimodal interaction}}. \bibinfo{pages}{216--220}.
\newblock


\bibitem[Vaswani et~al\mbox{.}(2017)]%
        {vaswani2017attention}
\bibfield{author}{\bibinfo{person}{Ashish Vaswani}, \bibinfo{person}{Noam Shazeer}, \bibinfo{person}{Niki Parmar}, \bibinfo{person}{Jakob Uszkoreit}, \bibinfo{person}{Llion Jones}, \bibinfo{person}{Aidan~N Gomez}, \bibinfo{person}{{\L}ukasz Kaiser}, {and} \bibinfo{person}{Illia Polosukhin}.} \bibinfo{year}{2017}\natexlab{}.
\newblock \showarticletitle{Attention is all you need}.
\newblock \bibinfo{journal}{\emph{Advances in neural information processing systems}}  \bibinfo{volume}{30} (\bibinfo{year}{2017}).
\newblock


\bibitem[Vinyals et~al\mbox{.}(2016)]%
        {vinyals2016matching}
\bibfield{author}{\bibinfo{person}{Oriol Vinyals}, \bibinfo{person}{Charles Blundell}, \bibinfo{person}{Timothy Lillicrap}, \bibinfo{person}{Daan Wierstra}, {et~al\mbox{.}}} \bibinfo{year}{2016}\natexlab{}.
\newblock \showarticletitle{Matching networks for one shot learning}.
\newblock \bibinfo{journal}{\emph{Advances in neural information processing systems}}  \bibinfo{volume}{29} (\bibinfo{year}{2016}).
\newblock


\bibitem[Wang et~al\mbox{.}(2020)]%
        {wang2020generalizing}
\bibfield{author}{\bibinfo{person}{Yaqing Wang}, \bibinfo{person}{Quanming Yao}, \bibinfo{person}{James~T Kwok}, {and} \bibinfo{person}{Lionel~M Ni}.} \bibinfo{year}{2020}\natexlab{}.
\newblock \showarticletitle{Generalizing from a few examples: A survey on few-shot learning}.
\newblock \bibinfo{journal}{\emph{ACM computing surveys (csur)}} \bibinfo{volume}{53}, \bibinfo{number}{3} (\bibinfo{year}{2020}), \bibinfo{pages}{1--34}.
\newblock


\bibitem[Wei et~al\mbox{.}(2020)]%
        {wei2020theoretical}
\bibfield{author}{\bibinfo{person}{Colin Wei}, \bibinfo{person}{Kendrick Shen}, \bibinfo{person}{Yining Chen}, {and} \bibinfo{person}{Tengyu Ma}.} \bibinfo{year}{2020}\natexlab{}.
\newblock \showarticletitle{Theoretical analysis of self-training with deep networks on unlabeled data}.
\newblock \bibinfo{journal}{\emph{arXiv preprint arXiv:2010.03622}} (\bibinfo{year}{2020}).
\newblock


\bibitem[Xie et~al\mbox{.}(2020)]%
        {xie2020self}
\bibfield{author}{\bibinfo{person}{Qizhe Xie}, \bibinfo{person}{Minh-Thang Luong}, \bibinfo{person}{Eduard Hovy}, {and} \bibinfo{person}{Quoc~V Le}.} \bibinfo{year}{2020}\natexlab{}.
\newblock \showarticletitle{Self-training with noisy student improves imagenet classification}. In \bibinfo{booktitle}{\emph{Proceedings of the IEEE/CVF conference on computer vision and pattern recognition}}. \bibinfo{pages}{10687--10698}.
\newblock


\bibitem[Xu et~al\mbox{.}(2021)]%
        {xu2021self}
\bibfield{author}{\bibinfo{person}{Qiantong Xu}, \bibinfo{person}{Alexei Baevski}, \bibinfo{person}{Tatiana Likhomanenko}, \bibinfo{person}{Paden Tomasello}, \bibinfo{person}{Alexis Conneau}, \bibinfo{person}{Ronan Collobert}, \bibinfo{person}{Gabriel Synnaeve}, {and} \bibinfo{person}{Michael Auli}.} \bibinfo{year}{2021}\natexlab{}.
\newblock \showarticletitle{Self-training and pre-training are complementary for speech recognition}. In \bibinfo{booktitle}{\emph{ICASSP 2021-2021 IEEE International Conference on Acoustics, Speech and Signal Processing (ICASSP)}}. IEEE, \bibinfo{pages}{3030--3034}.
\newblock


\bibitem[Yalniz et~al\mbox{.}(2019)]%
        {yalniz2019billion}
\bibfield{author}{\bibinfo{person}{I~Zeki Yalniz}, \bibinfo{person}{Herv{\'e} J{\'e}gou}, \bibinfo{person}{Kan Chen}, \bibinfo{person}{Manohar Paluri}, {and} \bibinfo{person}{Dhruv Mahajan}.} \bibinfo{year}{2019}\natexlab{}.
\newblock \showarticletitle{Billion-scale semi-supervised learning for image classification}.
\newblock \bibinfo{journal}{\emph{arXiv preprint arXiv:1905.00546}} (\bibinfo{year}{2019}).
\newblock


\bibitem[Zappi et~al\mbox{.}(2012)]%
        {zappi2012network}
\bibfield{author}{\bibinfo{person}{Piero Zappi}, \bibinfo{person}{Daniel Roggen}, \bibinfo{person}{Elisabetta Farella}, \bibinfo{person}{Gerhard Tr{\"o}ster}, {and} \bibinfo{person}{Luca Benini}.} \bibinfo{year}{2012}\natexlab{}.
\newblock \showarticletitle{Network-level power-performance trade-off in wearable activity recognition: A dynamic sensor selection approach}.
\newblock \bibinfo{journal}{\emph{ACM Transactions on Embedded Computing Systems (TECS)}} \bibinfo{volume}{11}, \bibinfo{number}{3} (\bibinfo{year}{2012}), \bibinfo{pages}{1--30}.
\newblock


\bibitem[Zhang and Sawchuk(2012)]%
        {zhang2012usc}
\bibfield{author}{\bibinfo{person}{Mi Zhang} {and} \bibinfo{person}{Alexander~A Sawchuk}.} \bibinfo{year}{2012}\natexlab{}.
\newblock \showarticletitle{USC-HAD: A daily activity dataset for ubiquitous activity recognition using wearable sensors}. In \bibinfo{booktitle}{\emph{Proceedings of the 2012 ACM conference on ubiquitous computing}}. \bibinfo{pages}{1036--1043}.
\newblock


\bibitem[Zoph et~al\mbox{.}(2020)]%
        {zoph2020learning}
\bibfield{author}{\bibinfo{person}{Barret Zoph}, \bibinfo{person}{Ekin~D Cubuk}, \bibinfo{person}{Golnaz Ghiasi}, \bibinfo{person}{Tsung-Yi Lin}, \bibinfo{person}{Jonathon Shlens}, {and} \bibinfo{person}{Quoc~V Le}.} \bibinfo{year}{2020}\natexlab{}.
\newblock \showarticletitle{Learning data augmentation strategies for object detection}. In \bibinfo{booktitle}{\emph{Computer Vision--ECCV 2020: 16th European Conference, Glasgow, UK, August 23--28, 2020, Proceedings, Part XXVII 16}}. Springer, \bibinfo{pages}{566--583}.
\newblock


\end{thebibliography}

\newpage
\appendix
\section{Appendix}
\subsection{Investigating Other Components of Cross-Domain HAR}

\subsubsection{Impact of source data augmentation}
\label{sec:source_data_augmentation}
As detailed in Sec.\ \ref{teacher_training}, we apply the following eight augmentation transformations (as proposed in \cite{um2017data} and utilized in \cite{saeed2019multi, tang2020exploring}) to the labeled source data before performing end-to-end Teacher model training:
\emph{(i)} adding random Gaussian noise; 
\emph{(ii)} scaling signal by a random factor; 
\emph{(iii)} applying a random 3D rotation; 
\emph{(iv)} reversing the time direction of input signal; 
\emph{(v)} negating input signal values; 
\emph{(vi)} warping the signal; 
\emph{(vii)} shuffling channels randomly; 
\emph{(viii)} randomly perturbing parts of time-series signals.
We analyze the necessity for such augmentation by comparing the performance of our method, Cross-Domain HAR when the Teacher model is trained with and without augmented source data.

\begin{figure*}[t]
    \centering
    \includegraphics[ width=1.0\linewidth]{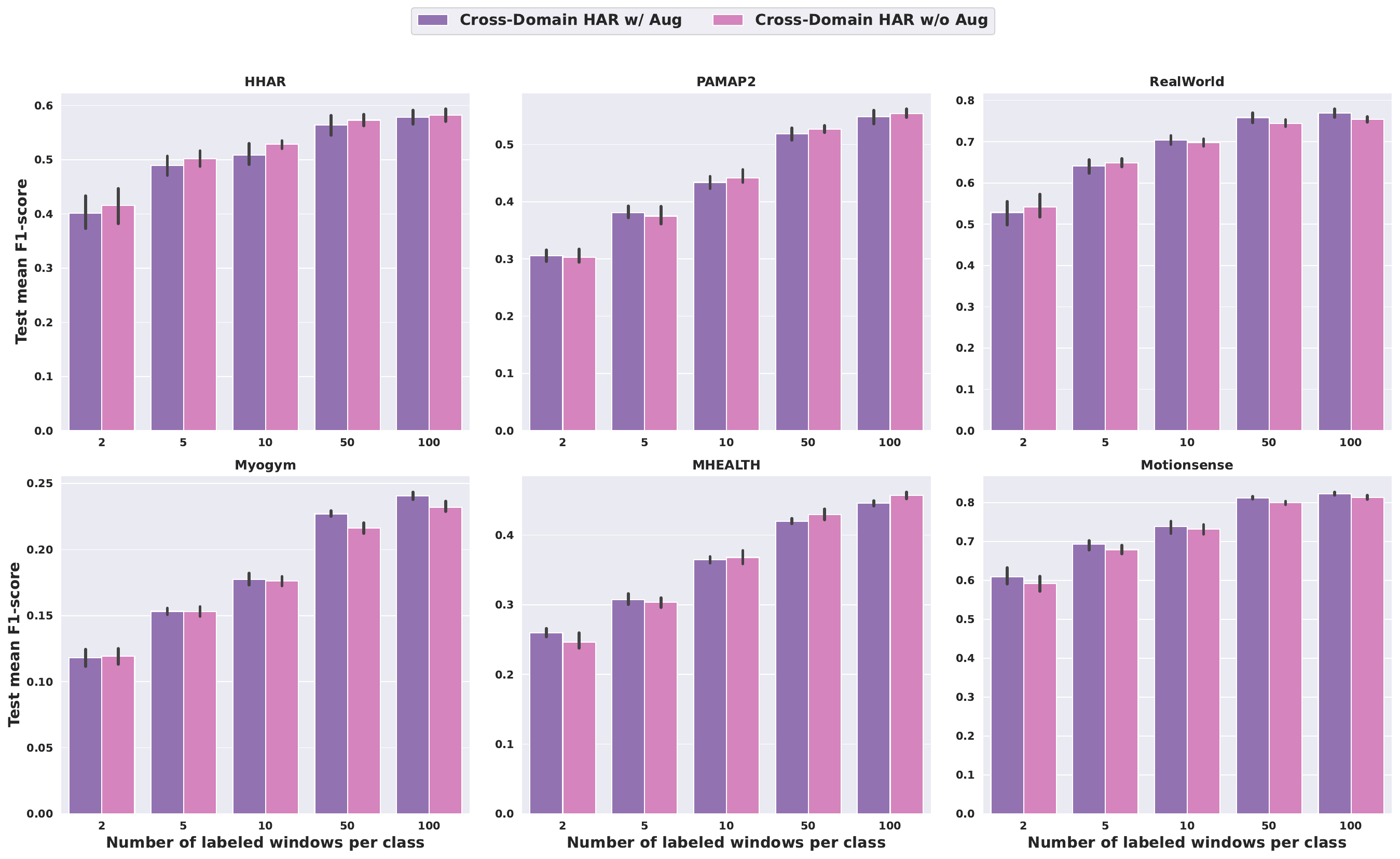}
    \caption{
    Analyzing the impact of using augmentations during Teacher model training (using Mobiact source dataset): 
    we study whether adding variation to the source data by applying eight augmentations \cite{um2017data, tang2020exploring} for Teacher model training is useful. 
    We  observe the positive impact of augmentations across three of six target datasets.  
    }
    \label{fig:smobi_noaug_results}
\end{figure*}

In Fig. \ref{fig:smobi_noaug_results}, for three of the six target datasets including Myogym, RealWorld and Motionsense, we observe that the performance obtained using augmented source data is higher by approx.\ 1-2\% .
For MHEALTH and Motionsense in particular, we observe slightly better performance by augmenting source data, when there is access to fewer (2 or 5) labeled windows per activity. 
These results support our hypothesis that adding variations to the source data assists in learning more generalized signal representations and achieving better recognition.

\begin{figure*}[!t]
    \centering
    \includegraphics[width=1.0\linewidth]{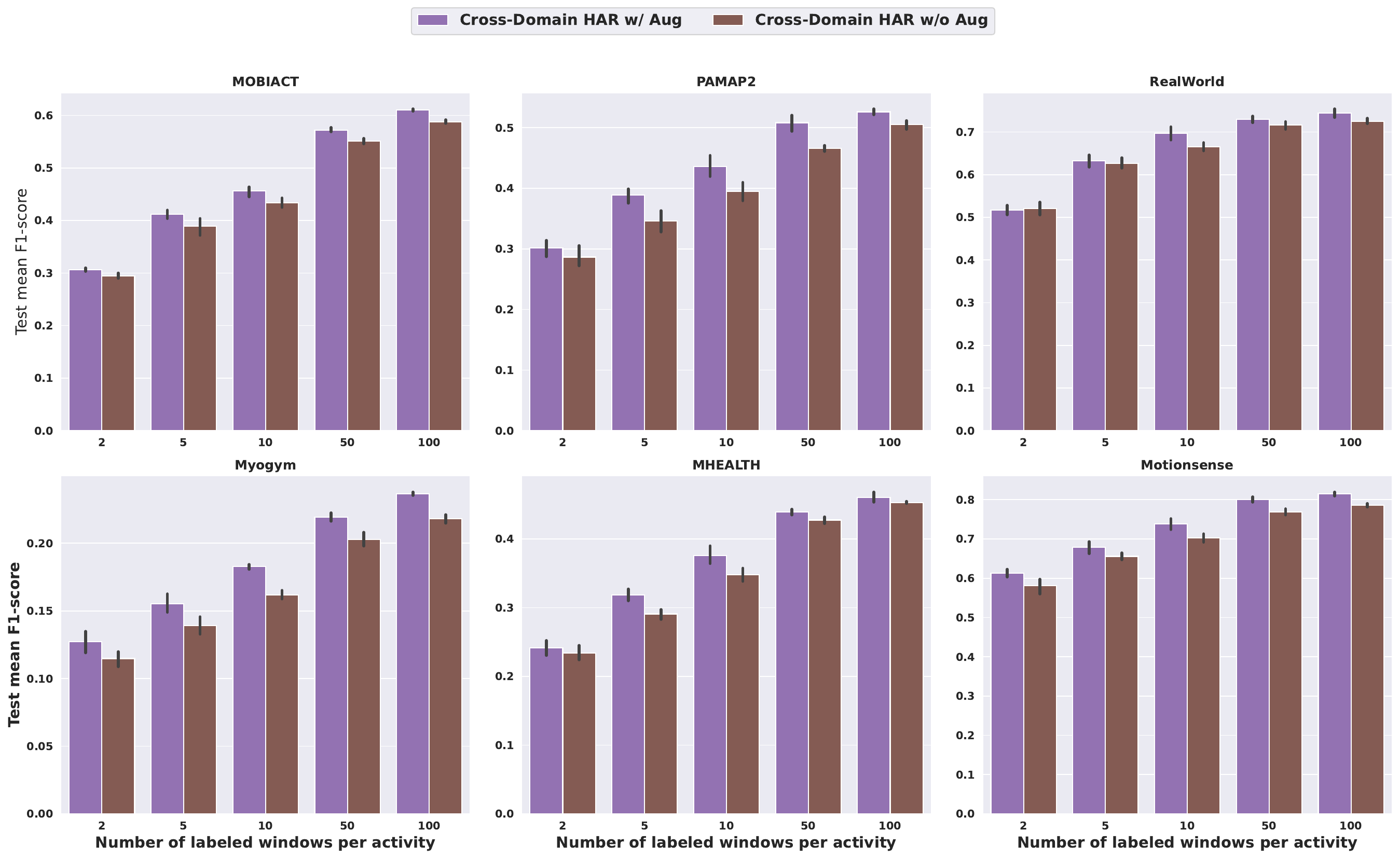}
    \caption{
    Analyzing the impact of using augmentations during Teacher model training (using HHAR source dataset): 
    we study whether adding variation to the HHAR source data by applying eight augmentations \cite{um2017data, tang2020exploring} for Teacher model training is useful. 
    We  observe the positive impact of augmentations across all target datasets.    }
    \label{fig:hhar_no_aug}
\end{figure*}

However, we note that when HHAR and PAMAP2 comprise the target datasets, augmenting source data has no performance benefits. 
We posit that data augmentation helps more in the case of smaller source datasets, and is not able to derive performance improvements for larger and more diverse source datasets such as Mobiact.
To assess this hypothesis and to analyze the impact of augmenting smaller source datasets, we study the performance on  HHAR in the Fig. \ref{fig:hhar_no_aug} where we observe the positive impact of augmentations across all target domains.

\subsubsection{Effect of using the consistency regularization during Student model training}
\label{sec:analysis_consist}

During self-training, we also perform consistency regularization by producing two different perturbations of the target unlabeled  data -- namely the weak and strong augmentations. 
The transformations applied for both weak and strong augmentations are detailed in Sec. \ref{pseudolabel_generation} and Sec. \ref{student_training} respectively.  
Here, we examine the impact of not generating weak and strong augmentations of target data, and instead calculating the KL-Divergence loss between teacher and student model predictions on original non-perturbed target data. 

\begin{figure*}[!t]
    \centering
    \includegraphics[ width=1.0\linewidth]{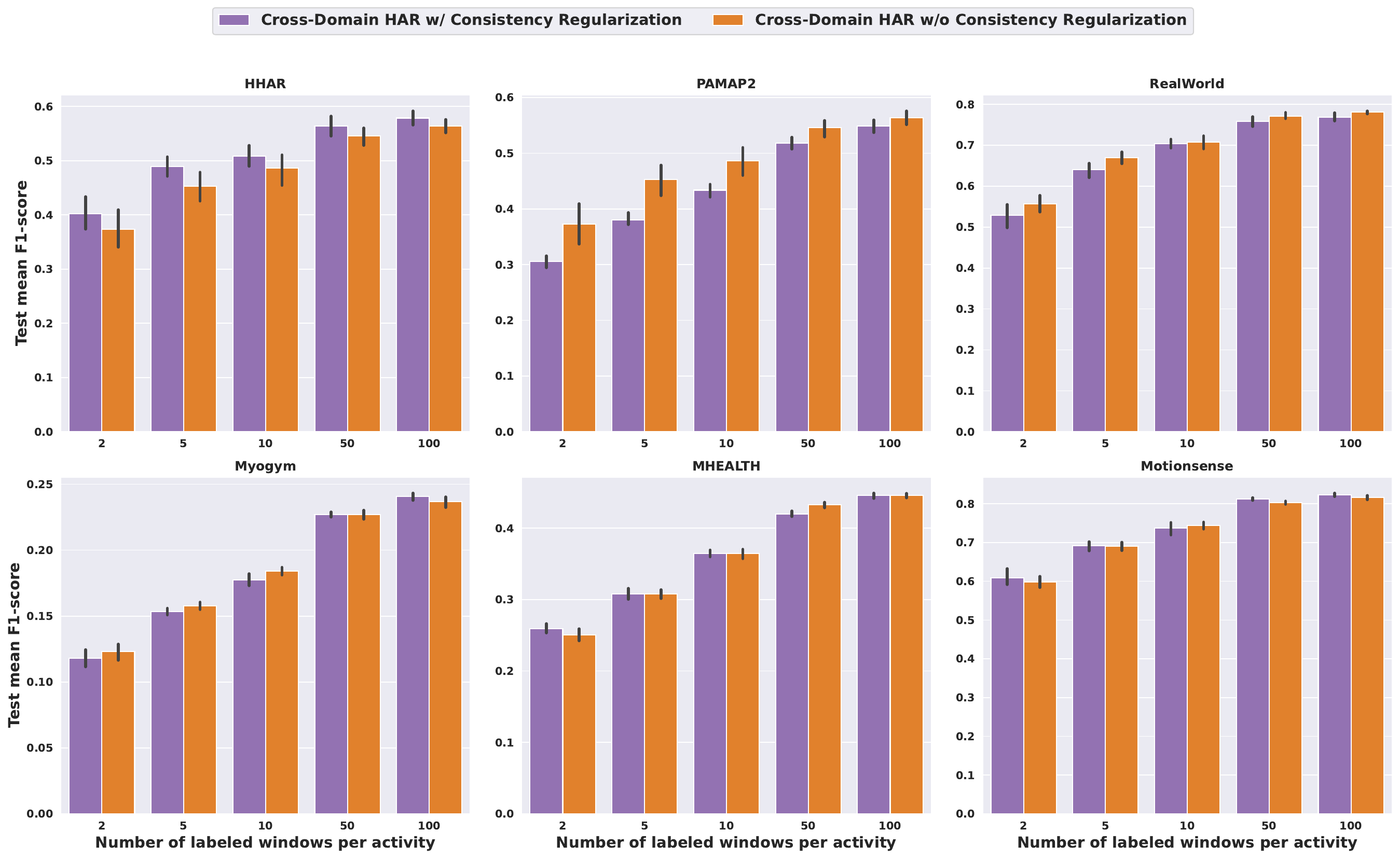}
    \caption{
    Effect of using consistency regularization during Student model training (using Mobiact source dataset):  
    we study whether the addition of consistency regularization has a positive impact on recognition, when utilized during self-training. 
    We observe that for two of six target datasets Cross-Domain w/ Consistency Regularization gives better performance. For rest of the target datasets, the performance is generally similar (PAMAP2 is an exception here.)
    }
    \label{fig:smobi_noconsist_results}
\end{figure*}

\begin{figure*}[!t]
    \centering
    \includegraphics[ width=1.0\linewidth]{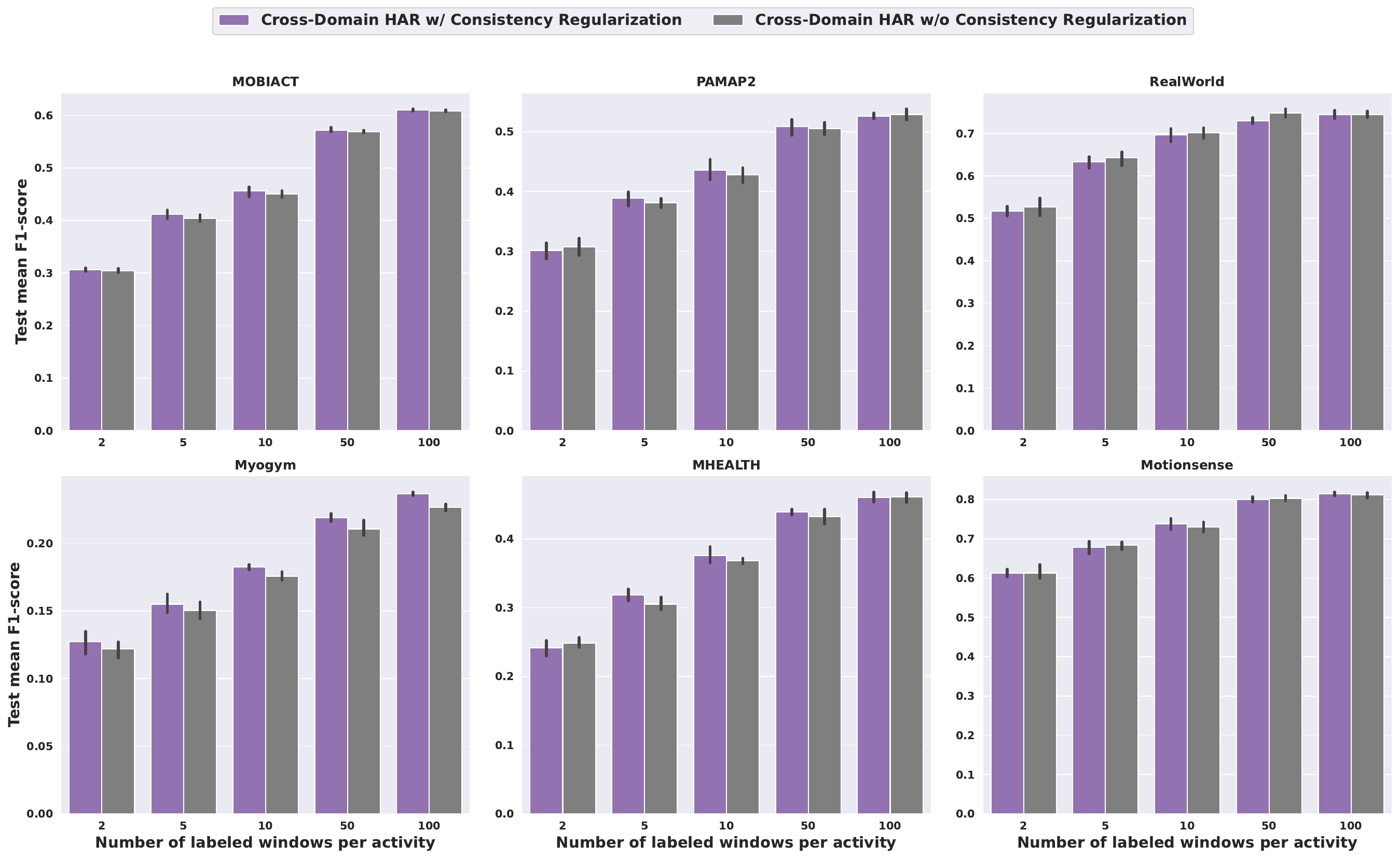}
    \caption{
    Effect of using consistency regularization during Student model training (using HHAR source dataset):  
    we study whether the addition of consistency regularization has a positive impact on recognition, when utilized during self-training especially for a smaller source dataset. 
    We observe that for four of six target datasets Cross-Domain w/ Consistency Regularization gives performance improvements.
    }
    \label{fig:shhar_noconsist_results}
\end{figure*}

In Fig. \ref{fig:smobi_noconsist_results}, we observe that the addition of consistency regularization leads to an increase in performance for HHAR, and in some cases for Myogym and Motionsense. 
The increase is around 2-3\% for HHAR dataset and around 1\% for Motionsense and Myogym (when >=50 annotated windows of activities are available). 
However, we observe when RealWorld and PAMAP2 are the target datasets, Student Model training without consistency regularization gives better results. 
We explore the impact of adding consistency regularization when utilizing a smaller source datasets, i.e., HHAR, in Fig. \ref{fig:shhar_noconsist_results}.
In this case, we observe clear performance improvements for most of target datasets (including Mobiact, PAMAP2, Myogym and  MHEALTH) demonstrating the positive impact of adding consistency-regularization to the self-training process.  

\begin{table}[h]
	\caption{Summary of the hyperparameter combinations used for our experimental evaluation.}

    \resizebox{\columnwidth}{!}{%
	\begin{tabular}{P{3.5cm}  P{10.5cm} }
        \toprule
	Method & Hyperparameter ranges \\ 
	\midrule
        CPC training w/ HHAR & Learning rates $\in \{1e-4, 1e-3, 5e-4 \}$, L2 reg. $\in  \{1e-4, 1e-5, 0 \}$, batch size $\in \{32,64, 128\}$ and number of future timesteps in $\in \{20,24,28,32\} $ \\
        \hline

        SimCLR training w/ HHAR and Mobiact & Learning rates $\in \{1e-4, 1e-3, 5e-4 \}$, L2 reg. $\in  \{1e-4, 1e-5, 0 \}$, batch size $\in \{1024, 2048, 4096\}$ \\
        \hline
        
         Student model training & Learning rates $\in \{1e-4, 1e-3, 5e-4 \}$, L2 reg. $\in  \{1e-4, 1e-5, 0 \}$, batch size $\in \{50, 128, 256, 512, 1024, 2048\}$, $\lambda_1, \lambda_2 \in \{0.1,0.2,0.3,0.4,0.5,0.6,0.7,0.8,0.9, 1.0\}$  \\ 
         
        \hline

        Few shot fine-tuning & Learning rates $\in \{1e-3, 1e-2, 1e-4, 5e-4, 5e-3, 8e-4\}$ and L2 reg. $\in  \{0, 1e-4, 1e-5, 1e-6\}$ \\
        \bottomrule

	\end{tabular}
 }
	\label{tab:hyperparam}

\end{table}


\begin{figure*}[!t]
    \centering
    \includegraphics[ width=1.0\linewidth]{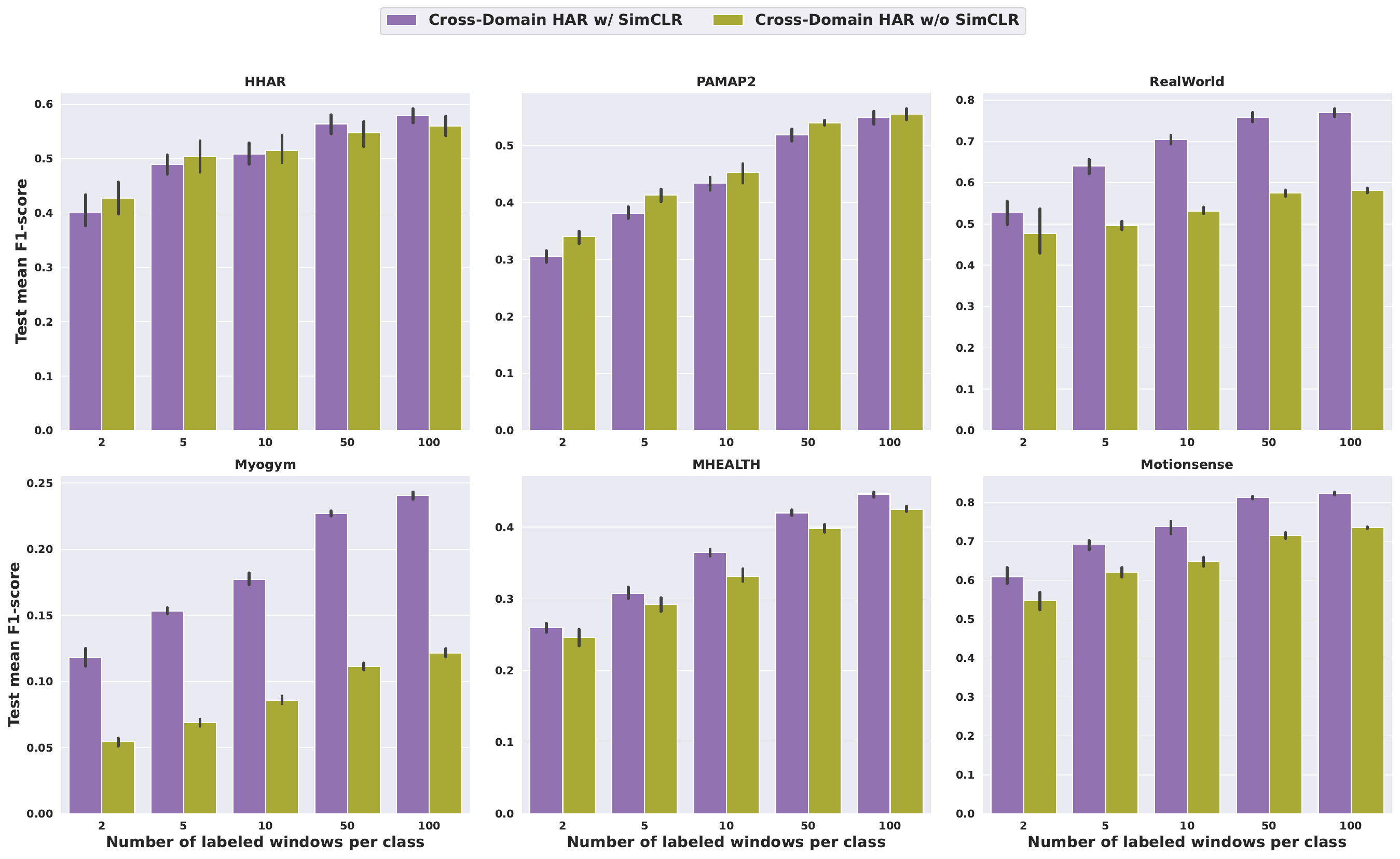}
    \caption{
    Effect of using self-supervised loss during Student model training (using Mobiact source dataset):  
    we study whether the addition of SimCLR-based self-supervision has a positive impact on recognition, when utilized during self-training. 
    We observe that Cross-Domain HAR w/o SimCLR performs  lower on most target datasets. 
    }
    \label{fig:smobi_noss_results}
\end{figure*}

\begin{figure*}[!t]
    \centering
    \includegraphics[width=1.0\linewidth]{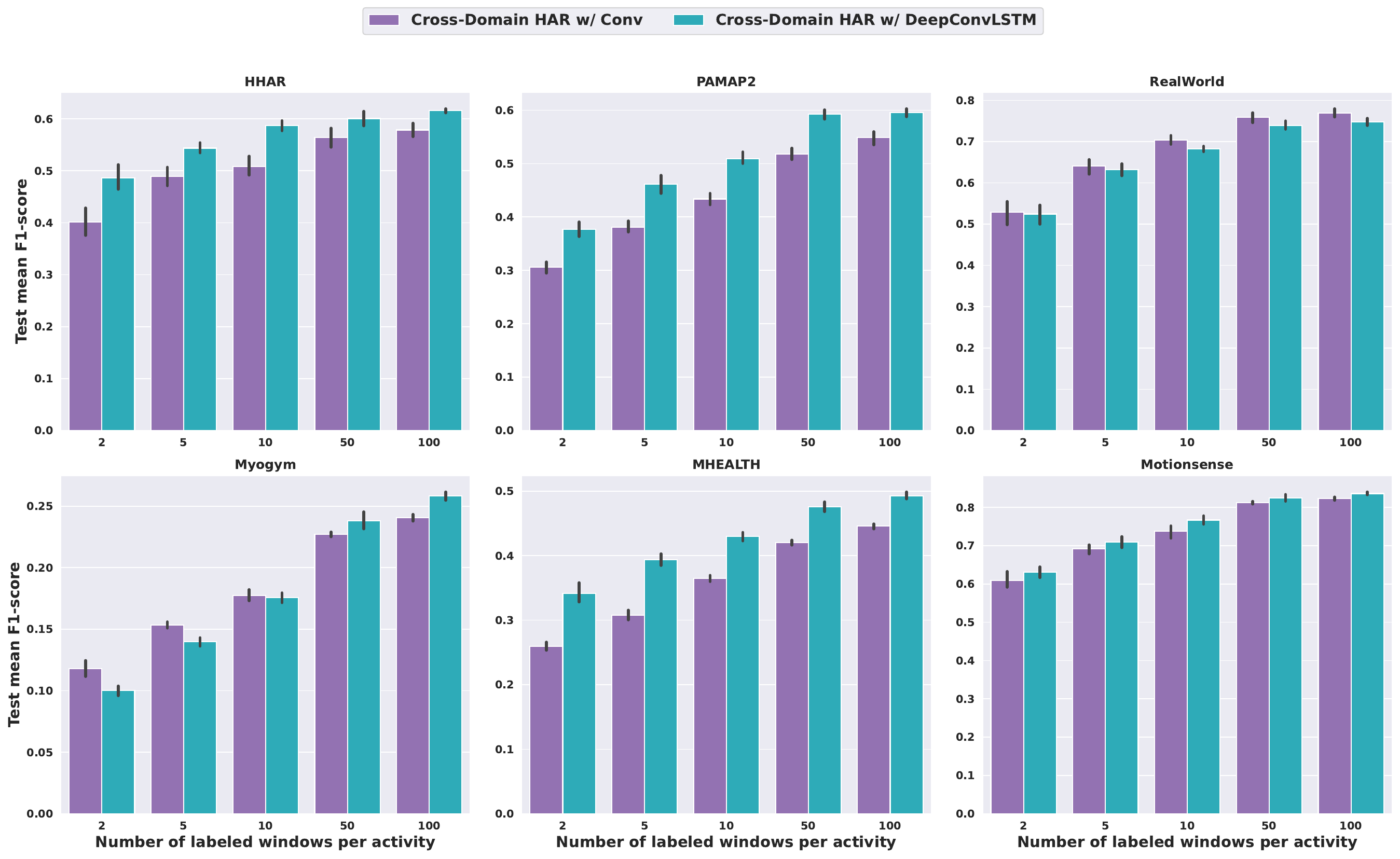}
    \caption{
    Impact of using different encoders in Cross-Domain HAR (using Mobiact source dataset): we replace the convolutional encoder with DeepConvLSTM \cite{ordonez2016deep}. 
    For four out of six target datasets including PAMAP2 and Motionsense, DeepConvLSTM performs better or comparable to convolutional encoder. 
    The choice of encoder has stronger impact when the number of labeled windows/class is lower, as seen for HHAR, PAMAP2 and  MHEALTH. 
    }
    \label{fig:convrec_results}
\end{figure*}

\end{document}